\documentclass{tmsce}

\usepackage{lipsum}
\usepackage{amsmath,amsfonts,amsthm}
\usepackage{algorithmic}
\usepackage{algorithm}
\usepackage{array}
\usepackage[caption=false,font=normalsize,labelfont=sf,textfont=sf]{subfig}
\usepackage{textcomp}
\usepackage{stfloats}
\usepackage{verbatim}
\usepackage{graphicx}
\usepackage{cite}
\usepackage{booktabs}
\usepackage{float}
\usepackage{placeins}
\usepackage{multirow}

\hypersetup{
    colorlinks=true,
    linkcolor=blue,
    citecolor=blue,
    urlcolor=blue
}

\doi{10.0000/tmsce.2026.001}

\copyrightline{© Published by
\textbf{Transactions on Mathematical Sciences and Computational Engineering}.\\
The Tribhuj Publishers}

\permissions{For permissions, please contact:\\
\url{tmsce.editorial@gmail.com}}

\vol{1}
\issue{1}
\yearofpub{2026}
\pagerange{}

\title{
HAMNO: A Hierarchical Adaptive Multi-scale Neural Operator
with Physics-Informed Learning for Dynamical Systems
}

\author{}
\date{}

\makeatletter
\renewcommand{\footnoterule}{%
  \kern 2.0em
  \hrule \@width 0.40\columnwidth \@height 0.4pt
  \kern 0.45em
}
\makeatother

\begin{document}

\makeatletter
\def\ps@plain{%
  \def\@oddhead{}%
  \def\@evenhead{}%
  \def\@oddfoot{\hfill\thepage\hfill}%
  \def\@evenfoot{\hfill\thepage\hfill}%
}
\pagestyle{plain}
\makeatother

\begin{center}

{\LARGE\bfseries
HAMNO: A Hierarchical Adaptive Multi-scale Neural Operator
with Physics-Informed Learning for Dynamical Systems\par}

\vspace{1em}

{\normalsize\bfseries Mostafa Bamdad$^{1,*}$, Mohammad Sadegh Eshaghi$^{2}$, Timon Rabczuk$^{1,\dagger}$\par}

\vspace{0.7em}
{\small $^{1}$Chair of Computational Mechanics, Institute of Structural Mechanics, Bauhaus-Universität Weimar, 99423 Weimar, Germany\par}
{\small $^{2}$Chair of Computational Science and Simulation Technology, Institute of Photonics, Department of Mathematics and Physics, Leibniz University Hannover, 30167 Hannover, Germany\par}

\end{center}

\begingroup
\renewcommand{\thefootnote}{\fnsymbol{footnote}}
\footnotetext[1]{Corresponding author: \href{mailto:mostafa.bamdad@uni-weimar.de}{mostafa.bamdad@uni-weimar.de}.}
\footnotetext[2]{Corresponding author: \href{mailto:timon.rabczuk@uni-weimar.de}{timon.rabczuk@uni-weimar.de}.}
\endgroup

\vspace{1.0em}

\begin{abstract}
Neural operators provide a powerful framework for learning solution mappings of partial differential equations directly in function space. However, many existing architectures still struggle to represent nonlinear time-dependent systems that involve multi-scale structures, long-range interactions, and stable long-time evolution. In this work, we introduce the Hierarchical Adaptive Multi-scale Neural Operator (HAMNO), a neural-operator architecture that combines local convolutional representations, global spectral operators, and hierarchical encoder--decoder processing. The central component of HAMNO is a data-dependent gating mechanism that adaptively balances local and global information at each spatial location, allowing the model to resolve fine-scale features while preserving long-range dependencies.

We further develop a physics-informed extension, PI-HAMNO, based on a multi-objective loss strategy that combines data fitting with strong- and weak-form physics constraints. The strong-form term penalizes the domain-integrated squared PDE residual in physical coordinates, while the weak-form term is constructed by multiplying the governing residual by finite-element test functions and evaluating the resulting element integrals using centroid-based tetrahedral quadrature. The framework is evaluated on non-periodic Allen--Cahn (AC), Cahn--Hilliard (CH), and Swift--Hohenberg (SH) equations defined on cubic domains. Across long-horizon rollout, data-limited training, out-of-distribution initial-condition shifts, and random-seed variations, HAMNO improves predictive accuracy over standard neural-operator baselines, while PI-HAMNO further enhances stability, physical consistency, and data efficiency. The implementation is publicly available at \url{https://github.com/MBamdad/HAMNO}.
\end{abstract}

\vspace{1.2em}


\printkeywords
\tmsceendfrontmatter

\section{Introduction}
\noindent

Many problems in computational mechanics, materials science, and pattern-forming systems are governed by nonlinear partial differential equations (PDEs) whose solutions evolve across several spatial and temporal scales. Classical numerical methods, including finite-element, finite-volume, finite-difference, and spectral schemes, remain the standard tools for solving such systems with high accuracy. However, repeated high-fidelity simulations can become prohibitively expensive when many trajectories, parameter values, initial conditions, or long-time predictions are required. Neural operators provide a direct route to this many-query setting by learning solution mappings in function space, allowing trained models to evaluate new input functions and discretizations at substantially lower cost~\cite{lu2021deeponet,lu2022fairno,kovachki2023parametric,aldirany2023green,patel2022variationally}.

In neural-operator learning, the goal is to learn the underlying solution operator rather than a single solution instance. This distinction is essential for time-dependent and parametric PDEs, where the input may represent an initial condition, coefficient field, forcing term, geometry, or previous solution history. DeepONet introduced a branch--trunk representation for nonlinear operators~\cite{lu2021deeponet}, while subsequent Fourier-, wavelet-, Laplace-, geometry-aware, and attention-based operators have broadened the range of applications to multi-scale dynamics, non-periodic signals, complex geometries, and high-dimensional solution fields~\cite{tran2023ffno,gupta2021multiwavelet,tripura2023wno,cao2024lno,li2022geofno,cao2021galerkin,li2022oformer,kontolati2024latent}. Nevertheless, accurate long-time prediction remains difficult when the dynamics contain localized interfaces, sharp transitions, non-dominant high-frequency modes, or strong coupling between local and global structures.

This difficulty is particularly relevant for Fourier-type neural operators. Although spectral layers efficiently represent long-range correlations, truncating spectral modes may weaken the recovery of localized and high-frequency features~\cite{tran2023ffno,loglofno2025}. LOGLO-FNO addresses this limitation by enriching Fourier operators with local spectral convolutions and high-frequency propagation mechanisms~\cite{loglofno2025}. However, such strategies mainly improve spectral reconstruction and do not fully resolve the broader challenge of adaptive local--global coupling, hierarchical multi-resolution representation, and physics-guided stability during long autoregressive rollout.

Several extensions have been proposed to address related limitations. Factorized Fourier Neural Operators (F-FNO)~\cite{tran2023ffno} improve spectral efficiency and flexibility. Wavelet and multiwavelet neural operators introduce localized basis representations that are more suitable for multi-scale solution fields~\cite{gupta2021multiwavelet,tripura2023wno}. Laplace neural operators improve the treatment of transient and non-periodic responses by learning in a transformed operator space~\cite{cao2024lno}. In parallel, U-shaped and encoder--decoder operator architectures such as U-FNO~\cite{wen2022ufno}, U-NO~\cite{rahman2022uno}, and U-Net~\cite{ronneberger2015unet} enhance multi-scale feature learning through hierarchical contraction--expansion paths and skip connections, leading to improved data efficiency and coarse-to-fine representation learning. However, these architectures primarily rely on fixed hierarchical feature fusion and local convolutional interactions, which can limit adaptive coupling between local and global dynamics across multiple spatial scales. Other developments focus on long-horizon temporal modeling. The Multi-Head Neural Operator (MHNO)~\cite{eshaghi2026mhno} introduces step-wise projection operators and temporal correction mechanisms to reduce rollout error accumulation, while recurrent neural-operator formulations have also been explored to improve long-time integration of dynamical systems~\cite{michalowska2024long}.

Recently, hybrid operator-learning paradigms have also been explored. Neural Operator Warm Starts (NOWS)~\cite{nows2025} combines neural operators with classical iterative solvers by generating high-quality initial guesses that accelerate convergence while preserving the robustness of traditional numerical methods. Physics-informed and variational operator-learning approaches, including VINO~\cite{eshaghi2025vino}, physics-informed wavelet neural operators~\cite{navaneeth2024piwno}, physics-informed latent neural operators~\cite{karumuri2025pilatent}, and variationally mimetic operator networks~\cite{patel2022variationally}, incorporate PDE structure, weak formulations, or energy-based constraints directly into the learning process to improve physical consistency and data efficiency. Hybrid frameworks such as PENCO~\cite{BAMDAD2026118862} further enhance long-time stability through numerical consistency and energy-based regularization. However, many of these formulations are primarily developed for periodic settings, simplified benchmark problems, or spectral discretizations, which makes their extension to three-dimensional non-periodic phase-field dynamics less straightforward.

Phase-field models provide a demanding setting for this class of methods because they combine nonlinear reaction terms, diffusion, fourth-order operators, conserved or non-conserved dynamics, and long-time morphological evolution. The AC equation describes non-conserved interfacial motion, the CH equation models conserved phase separation and coarsening, and the SH equation produces oscillatory pattern-forming structures. These systems are widely used in materials modeling and computational mechanics, but their accurate simulation remains computationally expensive, especially in three dimensions and under non-periodic boundary conditions~\cite{du2020phasefield,tourret2022phasefield,zhuang2022phasefield,montes2021accelerating,krischok2024fast}. Recent learning-based studies have begun to accelerate phase-field and mechanics-related simulations~\cite{ESHAGHI2025129119,eshaghi2026mhno,BAMDAD2026118862}, yet stable and accurate long-horizon prediction of three-dimensional non-periodic phase-field systems remains challenging.

U-FNO~\cite{wen2022ufno}, U-NO~\cite{rahman2022uno}, and U-Net~\cite{ronneberger2015unet} provide important but distinct multi-scale baselines. U-FNO combines Fourier layers with embedded U-Net modules; however, the local and spectral components are fused through fixed additive interactions. U-NO extends neural operators with a U-shaped multi-resolution structure, yet its operator blocks remain primarily spectral and are mainly driven by global Fourier mixing. In contrast, U-Net is a purely local convolutional encoder--decoder that is effective for extracting spatial features, but it does not explicitly model global operator interactions. Consequently, these architectures have limited ability to adaptively coordinate local spatial structures and global operator dynamics across multiple scales and evolving solution regimes.

To address this limitation, we introduce HAMNO, a new operator-learning framework that combines local convolutional operators and global spectral operators within a hierarchical encoder--decoder architecture. Unlike existing multi-scale neural operators that rely on fixed feature fusion mechanisms, HAMNO introduces a data-dependent gating strategy that adaptively balances local and global information at each spatial location. The hierarchical multi-scale structure further enables simultaneous learning of coarse global dynamics and fine-scale interfacial structures, improving long-horizon stability and representation capability for nonlinear time-dependent PDEs. Moreover, each HAMNO block employs a residual operator update that adds both the adaptively fused local--global feature and the subsequent channel-mixing transformation through residual connections. This design improves optimization stability and supports deeper multi-scale operator learning.

To further improve physical consistency and predictive stability, we introduce a physics-informed extension referred to as PI-HAMNO. In this framework, physics-based constraints are incorporated directly into the training objective through strong-form and weak-form PDE formulations within a multi-objective regularization strategy. By combining complementary numerical representations of the governing equations, the model captures different physical characteristics of the same system more effectively, improving accuracy, stability, and data efficiency. The strong-form term minimizes the integrated squared PDE residual in physical coordinates, while the weak-form term evaluates element-wise variational residuals using finite-element test functions and centroid-based tetrahedral quadrature.

The proposed framework is evaluated on three-dimensional phase-field systems with homogeneous Neumann boundary conditions, including the AC, CH, and SH equations. The numerical results demonstrate improved long-horizon accuracy, stability, and data efficiency compared with existing neural operator architectures, while the physics-informed formulation further enhances physical consistency and predictive robustness.

\section{Methodology}
\noindent

This section describes the neural operator frameworks used to model the non-periodic AC, CH, and SH dynamics. All models take a short temporal history of three-dimensional phase fields as input and learn the mapping to the next state. Denoting the input history length by $T_{\mathrm{in}}$, the learning task is
\begin{equation}
\mathcal{G}_{\theta}: \big(u^{n-T_{\mathrm{in}}+1},\ldots,u^{n-1},u^{n}\big) \mapsto u_{\theta}^{n+1},
\label{eq:operator_map}
\end{equation}
where $u^n$ denotes the solution field at the discrete time level $t_n$, and $n$ represents the current time-step index along the trajectory. During evaluation, the learned one-step map is applied autoregressively to reconstruct the full temporal evolution. The central challenge is to preserve the correct physical behavior while maintaining high predictive accuracy over long horizons.

We first present standard neural operator backbones, including the Fourier Neural Operator, the Factorized Fourier Neural Operator, DeepONet, and U-shaped neural operators. We then introduce the proposed Hierarchical Adaptive Multi-scale Neural Operator (HAMNO), followed by its physics-informed extension PI-HAMNO.

\subsection{Learning setting}
Let
\begin{equation}
\mathbf{U}^{n}_{in}
=
\left[
u^{n-T_{\mathrm{in}}+1},
\ldots,
u^{n-1},
u^{n}
\right]
\end{equation}
denote the input history at time level $n$. The backbone operator returns a one-step prediction of the next state,
\begin{equation}
u_{\theta}^{n+1}(\mathbf{x})
=
\mathcal{G}_{\theta}\!\left(\mathbf{U}^{n}_{in},\mathbf{x}\right),
\end{equation}
where $\mathcal{G}_{\theta}$ means the learned neural operator parameterized by ${\theta}$, $\mathbf{x}=(x,y,z)$ denotes the spatial coordinate and $u_{\theta}^{n+1}$ is the predicted approximation of the reference solution $u^{n+1}$.

The model is trained on temporal windows extracted from full trajectories and is evaluated by applying the learned one-step predictor autoregressively until the final time. Spatial coordinates are appended to the input so that the learned mapping is tied to the physical domain.

\subsection{Fourier neural operator}
The Fourier neural operator represents the evolution map through repeated global spectral mixing in the spatial coordinates. Given the input history $\mathbf{U}^{n}_{in}$ and the spatial coordinate field $\mathbf{x}$, the first step is a lifting map
\begin{equation}
v_0(\mathbf{x}) = P\big(\mathbf{U}^{n}_{in}(\mathbf{x}),\mathbf{x}\big),
\label{eq:fno_lift}
\end{equation}
where $P$ is a learned pointwise linear embedding into a latent feature space.

Each operator block then updates the latent feature by combining a Fourier integral operator with a local channel mixing operator,
\begin{equation}
v_{\ell+1}(\mathbf{x})
=
\sigma\!\left(
\mathcal{K}_{\ell}(v_\ell)(\mathbf{x}) + W_{\ell}v_\ell(\mathbf{x})
\right),
\qquad \ell = 0,\dots,L-1,
\label{eq:fno_block}
\end{equation}
where $\sigma$ is a nonlinear activation. The spectral operator $\mathcal{K}_{\ell}$ is defined in Fourier space by
\begin{equation}
\mathcal{F}\big(\mathcal{K}_{\ell}(v_\ell)\big)(\boldsymbol{\xi})
=
R_{\ell}(\boldsymbol{\xi})\,\mathcal{F}(v_\ell)(\boldsymbol{\xi}),
\label{eq:fno_spectral}
\end{equation}
with $R_{\ell}(\boldsymbol{\xi})$ a learned complex multiplier restricted to a finite set of retained low frequency modes. The term $W_\ell$ denotes a pointwise linear map in physical space.

After $L$ layers, the final solution is obtained by a single global projection,
\begin{equation}
u_\theta^{n+1}(\mathbf{x}) = Q\big(v_L(\mathbf{x})\big),
\label{eq:fno_proj}
\end{equation}
where $Q$ is a learned output head. Hence FNO uses one global latent representation and one final projection to produce the next state.

\subsection{Factorized Fourier neural operator}
The factorized Fourier neural operator preserves the spectral philosophy of FNO but replaces the fully coupled three dimensional Fourier kernel by a structured decomposition. Let $v_\ell$ denote the latent feature at layer $\ell$. Instead of learning one fully coupled spectral kernel, the factorized operator decomposes the update into directional and pairwise spectral components,
\begin{equation}
\mathcal{K}^{FFNO}_\ell(v_\ell)
=
\mathcal{K}^{x}_\ell(v_\ell)
+
\mathcal{K}^{y}_\ell(v_\ell)
+
\mathcal{K}^{z}_\ell(v_\ell)
+
\mathcal{K}^{xy}_\ell(v_\ell)
+
\mathcal{K}^{xz}_\ell(v_\ell)
+
\mathcal{K}^{yz}_\ell(v_\ell).
\label{eq:ffno_factorized}
\end{equation}

The one dimensional terms act separately along each coordinate direction, while the pairwise terms restore cross directional couplings. In the implemented model, these components are combined with learnable gates and an additional local convolutional path,
\begin{equation}
\widetilde{\mathcal{K}}^{FFNO}_\ell(v_\ell)
=
\sum_{m} \gamma_{\ell,m}\,\mathcal{K}^{m}_\ell(v_\ell)
+
\gamma_{\ell,loc}\,\mathcal{C}_{loc}(v_\ell),
\label{eq:ffno_gated}
\end{equation}
where $\gamma_{\ell,m}$ are normalized learnable coefficients and $\mathcal{C}_{loc}$ is a local three dimensional convolution. 
In practice, this local term plays a crucial role for non-periodic boundary conditions, as it complements the global spectral operator with boundary-aware local corrections, leading to significantly improved accuracy and data efficiency.

The layer update is then written as
\begin{equation}
v_{\ell+1}
=
v_\ell
+
\Psi_\ell\!\left(
\widetilde{\mathcal{K}}^{FFNO}_\ell(v_\ell)
\right),
\label{eq:ffno_block}
\end{equation}
where $\Psi_\ell$ is a channel mixing map realized by pointwise convolutions and nonlinear activation. This residual form improves stability.

As in FNO, the final output is produced by a single global projection,
\begin{equation}
u_\theta^{n+1}(\mathbf{x}) = Q\big(v_L(\mathbf{x})\big).
\label{eq:ffno_proj}
\end{equation}
Therefore, the main distinction from FNO is not the output mechanism, but the structured spectral factorization inside each operator layer.

\subsection{DeepONet}
DeepONet decomposes the learned operator into a branch component that encodes the input history and a trunk component that encodes the evaluation coordinates. In the present three dimensional setting, the branch network maps the input history field $\mathbf{U}^{n}_{in}$ to a latent feature field
\begin{equation}
b(\mathbf{x}) = \mathcal{G}_{\mathrm{br}}\big(\mathbf{U}^{n}_{in}\big)(\mathbf{x}),
\label{eq:deeponet_branch}
\end{equation}
where $\mathcal{G}_{\mathrm{br}}$ denotes the branch encoder and $b(\mathbf{x})$ is the resulting latent branch feature at location $\mathbf{x}$.

The trunk network acts on the spatial coordinate and produces two modulation fields,
\begin{equation}
(\gamma(\mathbf{x}),\eta(\mathbf{x})) = \mathcal{G}_{\mathrm{tr}}(\mathbf{x}),
\label{eq:deeponet_trunk}
\end{equation}
where $\mathcal{G}_{\mathrm{tr}}$ denotes the trunk map, $\gamma(\mathbf{x})$ is a multiplicative modulation field, and $\eta(\mathbf{x})$ is an additive modulation field.

In the implementation used here, the branch network is a convolutional encoder and the trunk network is a multilayer perceptron. The two parts are combined through feature-wise modulation,
\begin{equation}
\widetilde{b}(\mathbf{x}) = b(\mathbf{x}) \odot \big(1+\gamma(\mathbf{x})\big) + \eta(\mathbf{x}),
\label{eq:deeponet_film}
\end{equation}
where $\odot$ denotes componentwise multiplication and $\widetilde{b}(\mathbf{x})$ denotes the modulated latent feature after combining branch and trunk information.

The final output is then produced by a projection head,
\begin{equation}
u_\theta^{n+1}(\mathbf{x}) = Q\big(\widetilde{b}(\mathbf{x})\big),
\label{eq:deeponet_output}
\end{equation}
where $Q$ denotes the final projection operator.

This architecture separates two roles clearly. The branch network captures the dependence on the input field history, while the trunk network captures the dependence on spatial location. Their combination yields an operator representation that naturally distinguishes between the input function and the evaluation coordinates.

\subsection{U-shaped neural operator}
Several neural operator architectures extend the original FNO framework through hierarchical multi-resolution representations. These models combine encoder--decoder structures with spectral or convolutional operators in order to improve coarse-to-fine feature learning and multi-scale representation capability~\cite{wen2022ufno,rahman2022uno,ronneberger2015unet}.

\paragraph{U-FNO}
U-FNO augments the Fourier neural operator with embedded local U-Net pathways inside the operator blocks. Similar to FNO, the latent representation is first constructed through the lifting map in Eq.~\eqref{eq:fno_lift}. The latent feature is then updated through repeated U-Fourier layers,
\begin{equation}
v_{\ell+1}
=
\sigma\!\left(
\mathcal{K}_{\ell}(v_\ell)
+
\mathcal{U}_{\ell}(v_\ell)
+
W_{\ell}(v_\ell)
\right),
\end{equation}
where $\mathcal{K}_{\ell}$ denotes the global spectral integral operator, $\mathcal{U}_{\ell}$ is a U-Net-based local convolutional operator acting in physical space, and $W_{\ell}$ is a learnable pointwise linear operator. The U-Net branch improves local spatial representation and recovery of fine-scale structures, while the Fourier branch captures long-range interactions. These components are combined through fixed additive fusion inside each U-Fourier layer.

\paragraph{U-NO}
U-NO extends neural operators through a hierarchical U-shaped multi-resolution architecture. Starting from the lifted representation $v_0$, the model progressively constructs coarse latent representations through neural-operator blocks and resolution changes,
\begin{equation}
v^{(1)} = \mathcal{B}_1(v_0), \quad
v^{(2)} = \mathcal{B}_2(\downarrow v^{(1)}), \quad
\dots
\end{equation}
where each $\mathcal{B}_i$ denotes an operator block acting at a prescribed spatial resolution. The decoder path reconstructs the solution through upsampling and skip connections,
\begin{equation}
\hat{v}^{(i)}
=
\mathcal{R}_i\!\left(
\mathrm{Up}(\hat{v}^{(i+1)}) \oplus v^{(i)}
\right),
\end{equation}
where $\mathcal{R}_i$ denotes the decoder/refinement operator at resolution level $i$, $\mathrm{Up}(\cdot)$ denotes upsampling, and $\oplus$ represents channel-wise concatenation with the corresponding encoder feature. This allows the model to combine coarse global information with high-resolution features. Unlike standard FNO, U-NO explicitly learns representations across multiple spatial resolutions, but its information flow is governed by the prescribed U-shaped operator structure rather than by an adaptive local--global fusion mechanism.

\paragraph{U-Net}
U-Net is a purely convolutional encoder--decoder architecture that operates entirely in physical space. The encoder progressively extracts hierarchical local features through convolution and downsampling,
\begin{equation}
v^{(i+1)} = \mathcal{C}_i(\downarrow v^{(i)}),
\end{equation}
where $\mathcal{C}_i$ denotes local convolutional operators and $\downarrow$ represents spatial downsampling. The decoder reconstructs the solution through upsampling and skip connections,
\begin{equation}
\hat{v}^{(i)}
=
\mathcal{D}_i\!\left(
\mathrm{Up}(\hat{v}^{(i+1)}) \oplus v^{(i)}
\right),
\end{equation}
where $\mathcal{D}_i$ denotes the decoder convolutional operator at resolution level $i$. This structure enables efficient local multi-scale feature extraction and accurate recovery of fine spatial details. However, unlike neural operators with spectral components, U-Net does not explicitly model global operator interactions in frequency space.

Although these architectures improve multi-scale representation learning through hierarchical encoder--decoder structures, they typically rely on predefined interactions between local and global representations. This motivates the development of HAMNO, which introduces adaptive local--global operator coupling within a hierarchical neural operator framework.

\subsection{Hierarchical Adaptive Multi-scale Neural Operator (HAMNO)}

\begin{figure}[t]
\centering
\includegraphics[width=\linewidth]{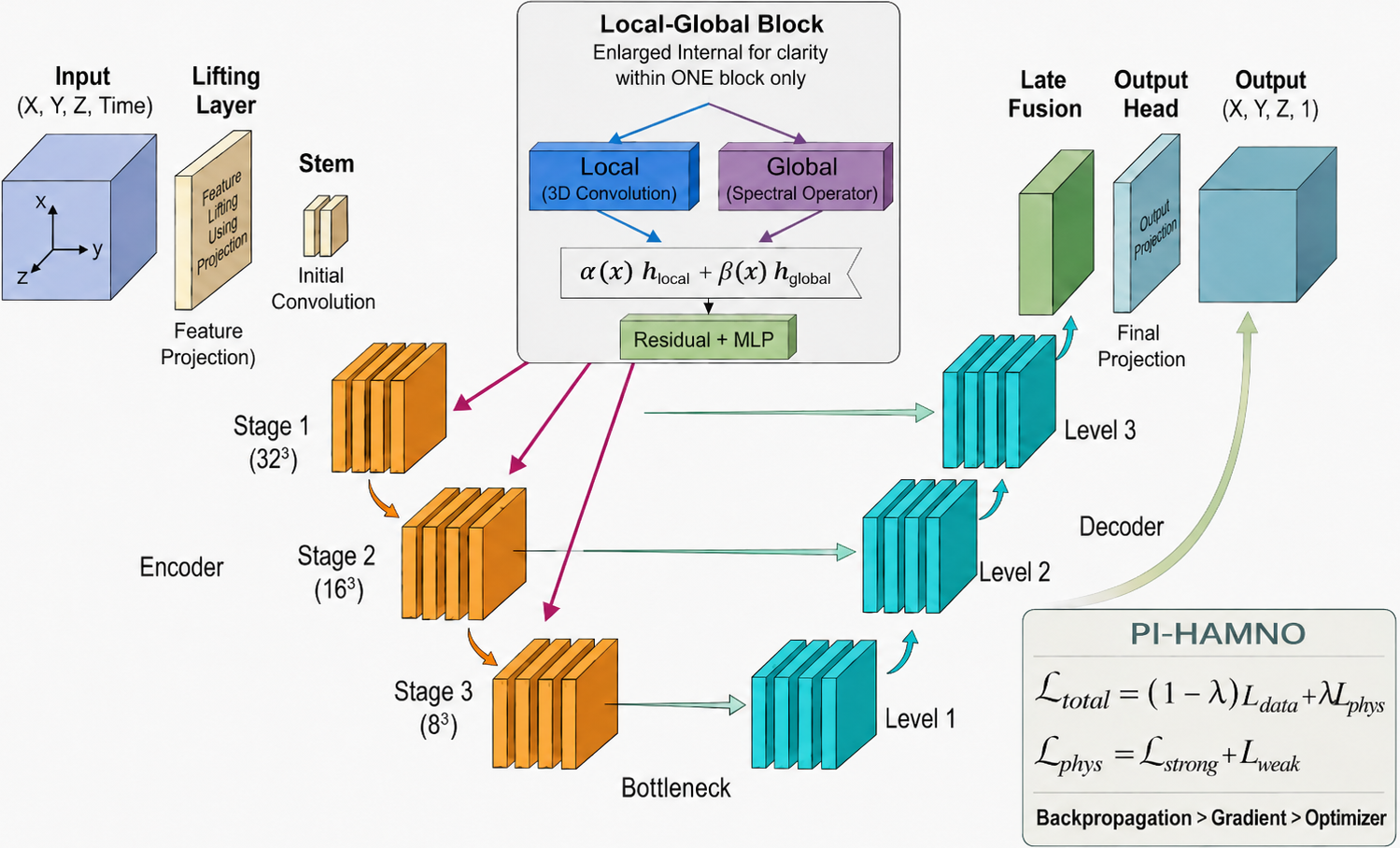}
\vspace{-26pt}
\caption{Schematic overview of HAMNO and its physics-informed extension PI-HAMNO, combining multi-scale operator learning with physics-based regularization.}
\label{fig:hamno}
\end{figure}

HAMNO is designed to represent nonlinear time-dependent dynamics by coupling local feature extraction, global spectral interaction, and hierarchical multi-scale processing within a single neural-operator architecture. A schematic overview of HAMNO and its physics-informed extension, PI-HAMNO, is shown in~\autoref{fig:hamno}.

As in the neural-operator backbones described above, the input history is first lifted into a latent feature space using the coordinate-aware embedding in Eq.~\eqref{eq:fno_lift}. The lifted representation is then evolved through adaptive local--global operator blocks arranged in a hierarchical encoder--decoder structure. In contrast to purely spectral operators or architectures with fixed local--global fusion, HAMNO learns a data-dependent balance between nearby spatial interactions and long-range spectral information at each resolution level.

\paragraph{Local-global operator decomposition}
At the core of HAMNO is a dual-path operator block, which serves as the fundamental building unit throughout all stages of the architecture. Given a latent feature representation $v_\ell(\mathbf{x})$, the model first applies normalization,
\begin{equation}
h(\mathbf{x}) = \mathrm{Norm}(v_\ell(\mathbf{x})).
\end{equation}

The normalized feature is then processed through two complementary operator branches. A local operator captures nearby spatial interactions,
\begin{equation}
h_{local} = \mathcal{K}_{local}(h),
\end{equation}
while a global operator models long-range dependencies through spectral transformation,
\begin{equation}
h_{global} = \mathcal{K}_{global}(h).
\end{equation}

Instead of combining these components through fixed additive fusion as in U-FNO, HAMNO introduces a data-dependent fusion mechanism,
\begin{equation}
h_{fused}(\mathbf{x})
=
\alpha(\mathbf{x})\, h_{local}(\mathbf{x})
+
\beta(\mathbf{x})\, h_{global}(\mathbf{x}),
\end{equation}
where the weights $\alpha(\mathbf{x})$ and $\beta(\mathbf{x})$ are learned adaptively from the input feature. This gating mechanism allows the model to dynamically balance local and global information depending on the spatial context and resolution scale.

The fused feature is then transformed as
\begin{equation}
\widetilde{h}_{fused} = \mathcal{M}(h_{fused}),
\end{equation}
where $\mathcal{M}$ denotes a $1\times1\times1$ convolution used to mix feature channels after adaptive local--global fusion. This operation prepares the fused representation for the residual update.

\paragraph{Residual operator update}
The block updates the latent feature through two residual steps. First, the mixed fused feature is added to the input feature,
\begin{equation}
v'_\ell = v_\ell + \widetilde{h}_{fused}.
\end{equation}
Then a nonlinear channel mixing map is applied through a second residual update,
\begin{equation}
v_{\ell+1} = v'_\ell + \Psi_\ell\!\left(\mathrm{Norm}(v'_\ell)\right),
\end{equation}
where $\Psi_\ell$ denotes a pointwise multilayer feature transformation. This structure improves stability and enables deeper operator learning.

\paragraph{Hierarchical multi-scale structure}
Unlike standard neural operators that operate at a single resolution, HAMNO adopts a hierarchical encoder--decoder structure to capture dynamics across multiple spatial scales. As illustrated in~\autoref{fig:hamno}, the encoder progressively compresses the spatial resolution (zooming out) to build increasingly abstract representations, while the decoder reconstructs the solution (zooming in) by recovering fine-scale details.

Starting from the lifted feature representation $v^{(1)}_0$, the model progressively constructs coarser representations through successive operator blocks and downsampling,
\begin{equation}
v^{(1)} = \mathcal{B}_1(v^{(1)}_0), \quad
v^{(2)} = \mathcal{B}_2(\downarrow v^{(1)}), \quad
v^{(3)} = \mathcal{B}_3(\downarrow v^{(2)}),
\end{equation}
where $\downarrow$ denotes spatial downsampling and each $\mathcal{B}_i$ represents a stack of local--global operator blocks. This process can be interpreted as progressively zooming out to capture large-scale and long-range interactions.

The representation is then reconstructed through an upsampling path with skip connections that preserve fine-scale information,
\begin{equation}
\hat{v}^{(2)} = \mathcal{R}_2\!\left(\mathcal{F}_2\big(\mathrm{Up}(v^{(3)}) \oplus v^{(2)}\big)\right),
\qquad
\hat{v}^{(1)} = \mathcal{R}_1\!\left(\mathcal{F}_1\big(\mathrm{Up}(\hat{v}^{(2)}) \oplus v^{(1)}\big)\right),
\end{equation}
where $\mathrm{Up}(\cdot)$ denotes learnable upsampling, $\oplus$ denotes channel-wise concatenation, $\mathcal{F}_1$ and $\mathcal{F}_2$ are learned fusion operators, and $\mathcal{R}_1$ and $\mathcal{R}_2$ denote refinement stages. The skip connections transfer high-resolution features from the encoder to the decoder, enabling accurate reconstruction by recovering details that would otherwise be lost during downsampling. This stage corresponds to zooming back in, guided by both coarse global context and preserved fine-scale information.

Finally, the reconstructed feature is fused with the early representation from the lifting stage,
\begin{equation}
v_{\text{final}} = \mathcal{F}_0\big(\hat{v}^{(1)} \oplus v^{(1)}_0\big),
\end{equation}
where $\mathcal{F}_0$ is a learned fusion operator implemented by $1 \times 1 \times 1$ convolutions followed by nonlinear activation. This hierarchical design enables the model to jointly learn global structures and fine-scale dynamics in a consistent manner.

\paragraph{Final prediction}
After multi-scale processing, the final output is obtained through a projection head,
\begin{equation}
u_\theta^{n+1}(\mathbf{x}) = Q\big(v_{\text{final}}(\mathbf{x})\big).
\end{equation}

This design enables HAMNO to simultaneously capture local dynamics, long-range interactions, and cross-scale dependencies within a unified operator framework.

\subsection{Physics-informed HAMNO (PI-HAMNO)}

To improve accuracy, stability, and data efficiency, we introduce a physics-informed extension of HAMNO, referred to as PI-HAMNO. In this framework, the neural operator is trained not only to fit available data, but also to satisfy numerical representations of the governing physical laws.

The total training objective is written as
\begin{equation}
\mathcal{L}_{\text{total}}
=
(1-\lambda)\mathcal{L}_{\text{data}}
+
\lambda \mathcal{L}_{\text{phys}},
\label{eq:pihamno_total_loss}
\end{equation}
where $\lambda \in [0,1]$ controls the balance between data fitting and physics-informed regularization.

Rather than relying on a single PDE formulation, we adopt a \emph{multi-objective regularization} strategy in which two complementary residual formulations of the same governing equation are combined:
\begin{equation}
\mathcal{L}_{\text{phys}}
=
\mathcal{L}_{\text{strong}}
+
\mathcal{L}_{\text{weak}}.
\label{eq:physics_multi_objective}
\end{equation}

The motivation for this construction is both physical and numerical. The strong-form residual measures the PDE defect directly in physical coordinates and therefore acts as an $L^2$-type penalty on local pointwise inconsistency. Since spatial derivatives amplify small-scale errors, this term is sensitive to sharp gradients, localized interfacial motion, and high-frequency residual components. It provides a direct local constraint on the learned time update, although it may also lead to a stiffer optimization landscape when used alone.

The weak-form residual introduces a complementary variational constraint. The governing residual is multiplied by finite-element test functions, integrated over tetrahedral elements, and, where appropriate, integration by parts is used to transfer derivatives from the solution to the test functions. This reduces the effective derivative order, naturally incorporates homogeneous Neumann boundary conditions, and evaluates the residual in an averaged element-wise sense rather than only through pointwise differential defects. As a result, the weak form emphasizes lower-frequency and globally consistent components of the dynamics and can be interpreted as a smoother variational measure of the PDE defect, closely related to a weaker dual norm.

Combining the two residuals therefore yields a principled physics regularization rather than an empirical addition of loss terms. The strong form constrains local and high-frequency inconsistencies, while the weak form improves variational consistency, conditioning, and global physical balance. This strong--weak coupling is also consistent with the design philosophy of HAMNO, where local and global information are learned jointly across multiple spatial scales. In this way, PI-HAMNO regularizes the learned dynamics through two complementary views of the same governing law:
\begin{itemize}
\item The strong-form residual enforces local PDE consistency by directly evaluating the residual using finite-difference approximations of spatial derivatives, followed by numerical integration of the squared residual over tetrahedral elements using centroid-based quadrature.
\item The weak-form residual enforces variational consistency by multiplying the governing residual by finite-element test functions, applying integration by parts to shift derivatives onto the test functions, and assembling element-wise residuals using shape functions, with integrals evaluated using tetrahedral quadrature.
\end{itemize}

\subsubsection{Strong-form residual}

Consider a PDE written abstractly as
\begin{equation}
u_t = \mathcal{N}(u),
\end{equation}
where $\mathcal{N}$ is a spatial differential operator. Given a previous state $u^n$ and a predicted state $u_\theta^{n+1}$, the time derivative is approximated by the one-step finite difference
\begin{equation}
u_t \approx \frac{u_\theta^{n+1}-u^n}{\Delta t}.
\end{equation}
The strong-form residual is then defined as
\begin{equation}
R_{\mathrm{FD}}
=
\frac{u_\theta^{n+1}-u^n}{\Delta t}
-
\mathcal{N}(u_\theta^{n+1}).
\end{equation}

In contrast to standard PINN formulations, where the residual is typically evaluated at sampled collocation points, we evaluate the residual on the structured grid and integrate its squared value over the physical domain. Spatial derivatives are approximated using finite differences. For example, the Laplacian is computed by the standard three-dimensional stencil
\begin{equation}
\Delta u_{i,j,k}
\approx
\frac{
u_{i+1,j,k}+u_{i-1,j,k}
+u_{i,j+1,k}+u_{i,j-1,k}
+u_{i,j,k+1}+u_{i,j,k-1}
-6u_{i,j,k}
}{(\Delta x)^2}.
\end{equation}
This local stencil makes the strong-form residual sensitive to pointwise differential imbalance, sharp gradients, and high-frequency errors.

To obtain a domain-integrated loss, each cubic grid cell is decomposed into six tetrahedra. For a tetrahedron $K$, the residual value at the centroid is approximated by averaging the residual values at its four vertices,
\begin{equation}
R_c^K
=
\frac{1}{4}
\sum_{a=1}^{4} R_a^K .
\end{equation}
The strong-form loss is then written as
\begin{equation}
\mathcal{L}_{\text{strong}}
\approx
\sum_{K}
\left(R_c^K\right)^2 |K|,
\end{equation}
where $|K|=(\Delta x)^3/6$ is the tetrahedral volume. Equivalently, this corresponds to the numerical approximation
\begin{equation}
\mathcal{L}_{\text{strong}}
\approx
\int_{\Omega}
\left(R_{\mathrm{FD}}\right)^2 d\mathbf{x}.
\end{equation}


\subsubsection{Weak-form residual}

To construct a weak-form residual, we start from the residual equation
\begin{equation}
u_t - \mathcal{N}(u) = 0,
\end{equation}
multiply it by a test function $v$, and integrate over the domain:
\begin{equation}
\int_\Omega \left(u_t - \mathcal{N}(u)\right)v\,d\Omega = 0.
\end{equation}
For operators containing second-order derivatives, integration by parts is applied to reduce the derivative order. This transfers derivatives from the solution field to the test functions and naturally incorporates homogeneous Neumann boundary conditions, for which the boundary contributions vanish.

The domain is partitioned into tetrahedral elements. On each element $K$, the solution is represented by linear P1 shape functions,
\begin{equation}
u^e(\mathbf{x})
=
\sum_{a=1}^{4} N_a(\mathbf{x}) u_a^e,
\end{equation}
where $u_a^e$ are the nodal values at the four vertices of the tetrahedron. The Galerkin choice is used for the test functions, i.e., $v=N_i$.

To define the element interpolation, we first introduce the reference tetrahedron $\widehat{K}$ with local coordinates $(\xi,\eta,\zeta)$. The P1 basis functions on $\widehat{K}$ are
\begin{align}
\widehat{N}_1 &= 1-\xi-\eta-\zeta, \\
\widehat{N}_2 &= \xi, \\
\widehat{N}_3 &= \eta, \\
\widehat{N}_4 &= \zeta .
\end{align}
These functions satisfy the nodal interpolation property and form a partition of unity. Each physical tetrahedron $K$ is obtained from $\widehat{K}$ through the affine mapping
\begin{equation}
\mathbf{x}
=
\mathbf{x}_1
+
J_K \widehat{\mathbf{x}},
\qquad
J_K
=
\left[
\mathbf{x}_2-\mathbf{x}_1,\,
\mathbf{x}_3-\mathbf{x}_1,\,
\mathbf{x}_4-\mathbf{x}_1
\right],
\end{equation}
where $\widehat{\mathbf{x}}=(\xi,\eta,\zeta)^T$ and $\mathbf{x}_a$ are the physical coordinates of the tetrahedral vertices. The physical shape functions are then obtained by composition with this mapping, so that
\begin{equation}
N_a(\mathbf{x})
=
\widehat{N}_a(\widehat{\mathbf{x}}(\mathbf{x})).
\end{equation}

Since the mapping is affine and the basis functions are linear, the gradients of the shape functions are constant inside each tetrahedron. Using the chain rule, the physical gradients are computed from the reference gradients as
\begin{equation}
\nabla N_a
=
J_K^{-T}\widehat{\nabla}\widehat{N}_a,
\qquad a=1,\ldots,4.
\end{equation}
For the reference P1 basis,
\begin{equation}
\widehat{\nabla}\widehat{N}_1=(-1,-1,-1)^T,\quad
\widehat{\nabla}\widehat{N}_2=(1,0,0)^T,\quad
\widehat{\nabla}\widehat{N}_3=(0,1,0)^T,\quad
\widehat{\nabla}\widehat{N}_4=(0,0,1)^T .
\end{equation}
Therefore, the element-wise gradient of the approximate solution is
\begin{equation}
\nabla u^e
=
\sum_{a=1}^{4} u_a^e \nabla N_a .
\end{equation}
The corresponding tetrahedral volume is
\begin{equation}
|K|
=
\frac{|\det J_K|}{6}.
\end{equation}

For each tetrahedron $K$ and local test function $N_i$, the weak residual has the generic form
\begin{equation}
r_i^K
=
\int_K u_t N_i\,dV
+
\int_K \text{(derivative terms)}\,dV
+
\int_K \text{(reaction terms)}\,dV .
\end{equation}
The exact structure of these terms depends on the governing equation. The derivative terms use the constant P1 gradients, while the time and nonlinear terms are evaluated by tetrahedral quadrature. In the centroid quadrature rule, the lower-order terms are evaluated at the tetrahedral centroid $\mathbf{x}_c$. Although the P1 shape functions vary linearly over the element, their values at the centroid are all equal to $1/4$:
\begin{equation}
N_i(\mathbf{x}_c)=\frac{1}{4},
\qquad i=1,\ldots,4.
\end{equation}
Therefore, terms of the form $\int_K f N_i\,dV$ are approximated as
\begin{equation}
\int_K f N_i\,dV
\approx
\frac{|K|}{4} f(\mathbf{x}_c),
\end{equation}
which gives an efficient element-wise assembly of the residual vector.

The weak-form loss is then defined by summing the squared residual entries over all tetrahedra and local test functions,
\begin{equation}
\mathcal{L}_{\text{weak}}
=
\frac{1}{N_e}
\sum_K \sum_i
\left(r_i^K\right)^2 .
\end{equation}
This formulation enforces the PDE in an integral sense, reduces the effective derivative order, improves numerical conditioning, and naturally accommodates Neumann boundary conditions.

\subsubsection{Specialization to the Allen--Cahn equation}

For the AC equation,
\begin{equation}
u_t
=
\Delta u
-
\frac{1}{\varepsilon^2}(u^3-u),
\end{equation}
the spatial operator is
\begin{equation}
\mathcal{N}_{\text{AC}}(u)
=
\Delta u
-
\frac{1}{\varepsilon^2}(u^3-u).
\end{equation}
The strong-form residual follows directly from the general formulation by evaluating the finite-difference Laplacian and nonlinear reaction term at the predicted state $u_\theta^{n+1}$.

For the weak form, inserting the AC operator into the Galerkin formulation gives the element residual
\begin{equation}
r_{i,\text{AC}}^K
=
\int_K u_t N_i\,dV
+
\int_K \nabla u \cdot \nabla N_i\,dV
+
\frac{1}{\varepsilon^2}
\int_K (u^3-u)N_i\,dV.
\end{equation}
The second term results from integration by parts of the diffusion operator, while the homogeneous Neumann boundary contribution vanishes naturally. The corresponding weak-form loss is
\begin{equation}
\mathcal{L}_{\text{weak}}^{\text{AC}}
=
\frac{1}{N_e}
\sum_K \sum_i
\left(r_{i,\text{AC}}^K\right)^2.
\end{equation}


\subsubsection{Specialization to the Cahn--Hilliard equation}

For the CH equation, the fourth-order dynamics are written in mixed second-order form as
\begin{equation}
u_t = \Delta \mu,
\qquad
\mu = -\varepsilon^2 \Delta u + (u^3-u),
\end{equation}
where $\mu$ denotes the chemical potential. This mixed representation is consistent with the finite-element weak formulation described above, since it avoids direct discretization of fourth-order derivatives.

The strong-form residual consists of two coupled residuals: one for the evolution equation and one for the chemical-potential relation. It is written as
\begin{equation}
\mathcal{L}_{\text{strong}}^{\text{CH}}
=
\int_{\Omega}
\left(
R_u^2
+
R_{\mu}^2
\right)
\,d\mathbf{x},
\end{equation}
where $R_u$ and $R_{\mu}$ denote the residuals associated with $u_t=\Delta\mu$ and $\mu=-\varepsilon^2\Delta u+(u^3-u)$, respectively.

The weak formulation produces the corresponding two element residual blocks. The evolution equation gives
\begin{equation}
r_{u,i}^K
=
\int_K u_t N_i\,dV
+
\int_K \nabla \mu \cdot \nabla N_i\,dV,
\end{equation}
and the chemical-potential relation gives
\begin{equation}
r_{\mu,i}^K
=
\int_K \mu N_i\,dV
-
\varepsilon^2
\int_K \nabla u \cdot \nabla N_i\,dV
-
\int_K (u^3-u)N_i\,dV.
\end{equation}
The corresponding weak-form loss is
\begin{equation}
\mathcal{L}_{\text{weak}}^{\text{CH}}
=
\frac{1}{N_e}
\sum_K \sum_i
\left[
\left(r_{u,i}^K\right)^2
+
\left(r_{\mu,i}^K\right)^2
\right].
\end{equation}
Thus, the CH formulation preserves the same tetrahedral P1 element structure as AC, while introducing coupled residuals for the conserved variable $u$ and the chemical potential $\mu$.

\subsubsection{Specialization to the Swift--Hohenberg equation}

For the SH equation,
\begin{equation}
u_t
=
\varepsilon u
-
u^3
-
(1+\Delta)^2 u,
\end{equation}
we again use a mixed second-order representation to avoid direct treatment of the fourth-order operator. Introducing
\begin{equation}
p = \Delta u,
\end{equation}
the evolution equation can be written as
\begin{equation}
u_t
=
(\varepsilon-1)u
-
u^3
-
2p
-
\Delta p.
\end{equation}

The primary strong-form residual is therefore
\begin{equation}
R_u
=
\frac{u_\theta^{n+1}-u^n}{\Delta t}
-
\left[
(\varepsilon-1)u
-
u^3
-
2p
-
\Delta p
\right].
\end{equation}
Here, the right-hand side is evaluated using the predicted state, or the corresponding time-collocation state when a midpoint evaluation is used.

Unlike the CH chemical potential, which is directly coupled to the evolution equation, the SH auxiliary variable $p=\Delta u$ mainly serves to split the fourth-order operator. To make this auxiliary variable physically informative during training, we introduce an additional consistency condition derived from the evolution equation. Specifically, two representations of $p$ are constructed:
\begin{align}
p_{\text{aux}} &= \Delta u, \\
(2+\Delta)p_{\text{evol}} &= (\varepsilon-1)u - u^3 - u_t.
\end{align}
The second relation follows directly from rearranging the SH evolution equation. The auxiliary residual is then defined as
\begin{equation}
R_p = p_{\text{aux}} - p_{\text{evol}}.
\end{equation}
In practice, the Helmholtz-type equation $(2+\Delta)p_{\text{evol}}=g$ is solved using a cosine spectral representation consistent with homogeneous Neumann boundary conditions, which provides a stable frequency-space inversion.

The total strong-form loss for SH is
\begin{equation}
\mathcal{L}_{\text{strong}}^{\text{SH}}
=
\int_{\Omega}
\left(R_u^2 + R_p^2\right)
\,d\mathbf{x}.
\end{equation}

The weak formulation follows the same tetrahedral P1 Galerkin framework used for AC and CH. The evolution equation gives
\begin{equation}
r_{u,i}^K
=
\int_K u_t N_i\,dV
-
\int_K \left[(\varepsilon-1)u - u^3 - 2p\right] N_i\,dV
-
\int_K \nabla p \cdot \nabla N_i\,dV.
\end{equation}
The auxiliary relation $p=\Delta u$ is written in weak form as
\begin{equation}
r_{p,i}^K
=
\int_K p N_i\,dV
+
\int_K \nabla u \cdot \nabla N_i\,dV,
\end{equation}
where integration by parts has been used. The weak-form loss is therefore
\begin{equation}
\mathcal{L}_{\text{weak}}^{\text{SH}}
=
\frac{1}{N_e}
\sum_K \sum_i
\left[
\left(r_{u,i}^K\right)^2
+
\left(r_{p,i}^K\right)^2
\right].
\end{equation}
Thus, the SH formulation uses the same numerical ingredients as the AC and CH cases: finite-difference strong residuals, tetrahedral P1 weak residuals, quadrature-based element integration, and strong--weak multi-objective physics regularization.

\section{Training and evaluation protocol}
\noindent
The numerical datasets used for training and testing are generated using the DCT-based non-periodic phase-field solvers described in Appendix~\ref{app:dataset_generation}. In brief, the reference trajectories correspond to the three-dimensional AC, CH, and SH benchmarks on cubic domains with homogeneous Neumann boundary conditions, for which the Laplacian is diagonalized using cosine modes. The resulting high-fidelity solution sequences are subsequently converted into temporal windows for one-step neural-operator training and long-horizon autoregressive evaluation.

Each trajectory is initialized from a Gaussian-random-field-based initial condition, producing a diverse set of phase-field morphologies with different interfacial patterns and characteristic length scales. The raw simulation outputs are converted into MATLAB \texttt{.mat} files and then arranged into temporal windows for one-step neural-operator training. In all cases, the models receive a short history of previous states and are trained to predict the next state, after which long-horizon trajectories are reconstructed by autoregressive rollout. The data-generation and learning workflows are executed on an NVIDIA A100 GPU with 40GB memory, using consistent numerical precision during preprocessing, training, and inference.

To ensure a fair comparison across models and training regimes, the evaluation protocol uses a fixed held-out test set for each benchmark, so that all models are evaluated on the same trajectories. Training settings, including the number of epochs, batch size, optimizer, and learning-rate schedule, are kept consistent within each benchmark and training configuration. In the pure-physics setting $(\lambda=1)$, the number of optimizer updates is kept independent of the nominal training-set size, preventing larger datasets from receiving additional gradient-update steps in this setting. Further implementation details are provided in~\cite{BAMDAD2026118862}.

\section{Numerical analysis}
\noindent
This section evaluates HAMNO and PI-HAMNO on the AC, CH, and SH benchmarks with homogeneous Neumann boundary conditions, focusing on predictive accuracy, training behavior, physical consistency, out-of-distribution generalization, and seed robustness. We first report in-distribution (ID) results using the original train--test data split and the fixed random seed $42$, including long-horizon rollout accuracy, training convergence, computational cost, and the effect of the physics weight $\lambda$. We then assess out-of-distribution (OOD) performance by evaluating the saved trained models on trajectories generated from unseen initial-condition parameter regimes. Finally, we examine robustness with respect to random initialization and data splitting by repeating selected experiments over multiple random seeds. The neural-network hyperparameters used for all models are reported in the supplementary information,~\autoref{tab:nn_hyperparameters}.

\subsection{In-distribution rollout accuracy and physical consistency}

Unless otherwise stated, the results in this subsection correspond to the in-distribution test setting with fixed random seed $42$. For each equation, the dataset contains $600$ total trajectories, with training subsets of $50$, $100$, or $200$ samples and a fixed held-out test set of $50$ samples.

\paragraph{Evaluation metric.}
The main accuracy measure is the relative $L^2$ error,
\begin{equation}
\mathrm{Rel}\text{-}L^2
=
\frac{\|u^{\mathrm{pred}}-u^{\mathrm{true}}\|_{L^2}}
{\|u^{\mathrm{true}}\|_{L^2}}.
\end{equation}
This metric is reported during autoregressive rollout, where each one-step prediction is fed back as input for the next step.

\begin{figure*}[!t]
\vspace*{-2.5em}
\centering

\makebox[\textwidth][c]{%
\begin{minipage}[c]{0.025\textwidth}
\centering
\textbf{(a)}
\end{minipage}
\hspace{-0.002\textwidth}
\begin{minipage}[c]{1.075\textwidth}
\centering
\includegraphics[width=\linewidth]{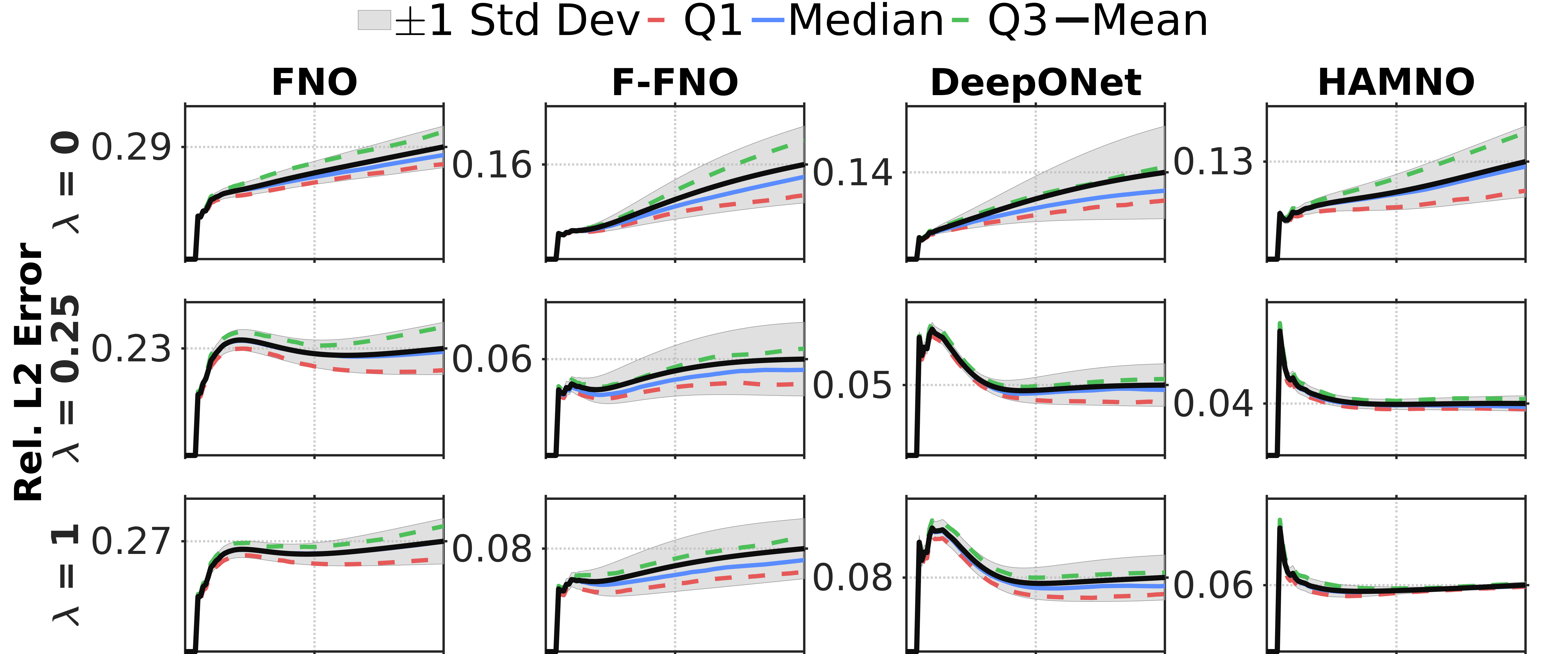}
\end{minipage}%
}

\vspace{0.32em}

\makebox[\textwidth][c]{%
\begin{minipage}[c]{0.025\textwidth}
\centering
\textbf{(b)}
\end{minipage}
\hspace{-0.002\textwidth}
\begin{minipage}[c]{1.075\textwidth}
\centering
\includegraphics[width=\linewidth]{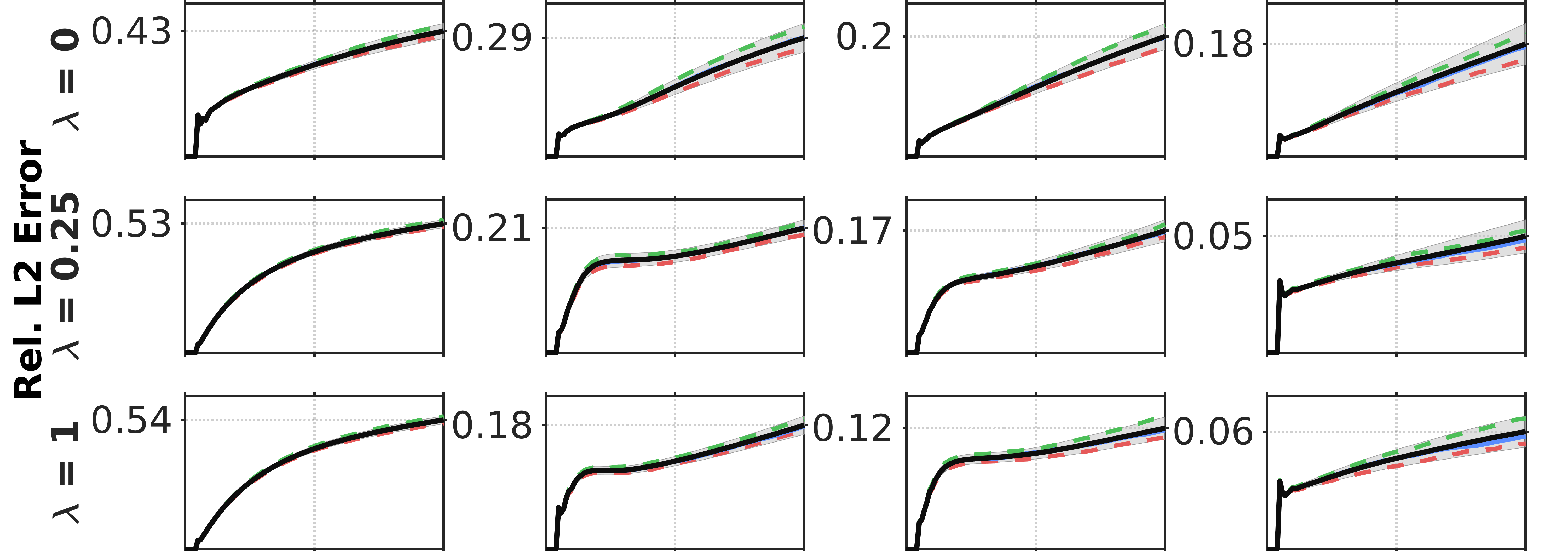}
\end{minipage}%
}

\vspace{0.32em}

\makebox[\textwidth][c]{%
\begin{minipage}[c]{0.025\textwidth}
\centering
\textbf{(c)}
\end{minipage}
\hspace{-0.002\textwidth}
\begin{minipage}[c]{1.075\textwidth}
\centering
\includegraphics[width=\linewidth]{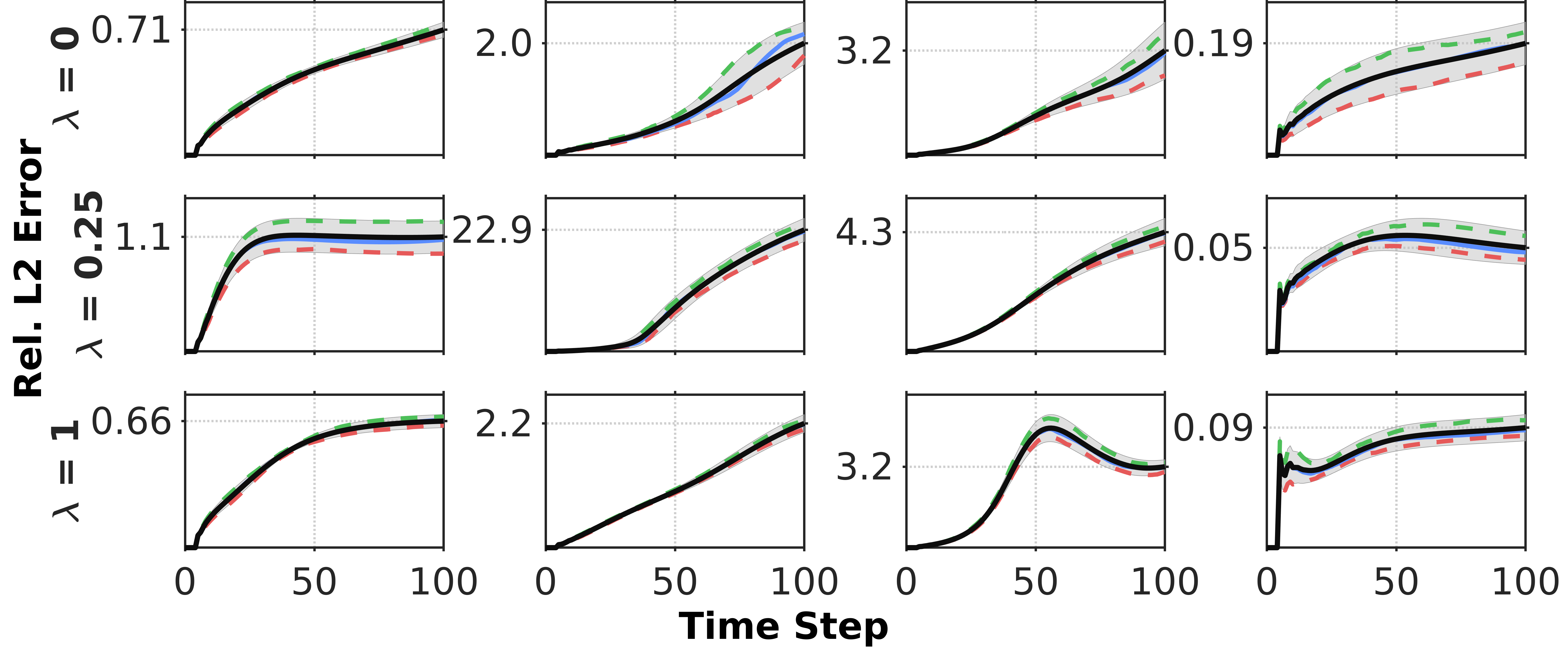}
\end{minipage}%
}

\vspace{-0.25em}

\caption{Rollout relative $L^2$ errors for the full neural-operator comparison on the three-dimensional phase-field benchmarks: (a) AC, (b) CH, and (c) SH. The vertically stacked layout compares long-horizon predictive behavior across the governing equations while preserving readability.}
\label{fig:compact_results}
\end{figure*}

The global rollout trends are summarized in~\autoref{fig:compact_results}. Across the AC, CH, and SH benchmarks, HAMNO consistently controls the accumulation of rollout error more effectively than the standard operator baselines, particularly at later time steps where small one-step inaccuracies are amplified by autoregressive prediction. This behavior reflects the main architectural advantage of HAMNO: the model does not rely only on global spectral mixing, as in FNO-type architectures, or only on local hierarchical features, as in convolutional encoder--decoders; instead, it adaptively combines local and global information at each spatial location and across multiple resolutions.

The benefit of this adaptive local--global coupling becomes more pronounced as the dynamics become stiffer and more multi-scale. For the CH and SH equations, where interfacial motion, fourth-order operators, and pattern-forming structures make long-time prediction more sensitive to phase errors, the physics-informed version PI-HAMNO further reduces error growth by guiding the learned dynamics toward the governing PDE structure. The corresponding loss curves show that the strong--weak residual regularization improves physical consistency without destabilizing training, indicating that the additional physics terms act as useful constraints rather than as competing objectives.

\begin{figure*}[!t]
\vspace*{-2.5em}
\centering

\makebox[\textwidth][c]{%
\begin{minipage}[c]{0.025\textwidth}
\centering
\textbf{(a)}
\end{minipage}
\hspace{-0.002\textwidth}
\begin{minipage}[c]{1.025\textwidth}
\centering
\includegraphics[width=\linewidth]{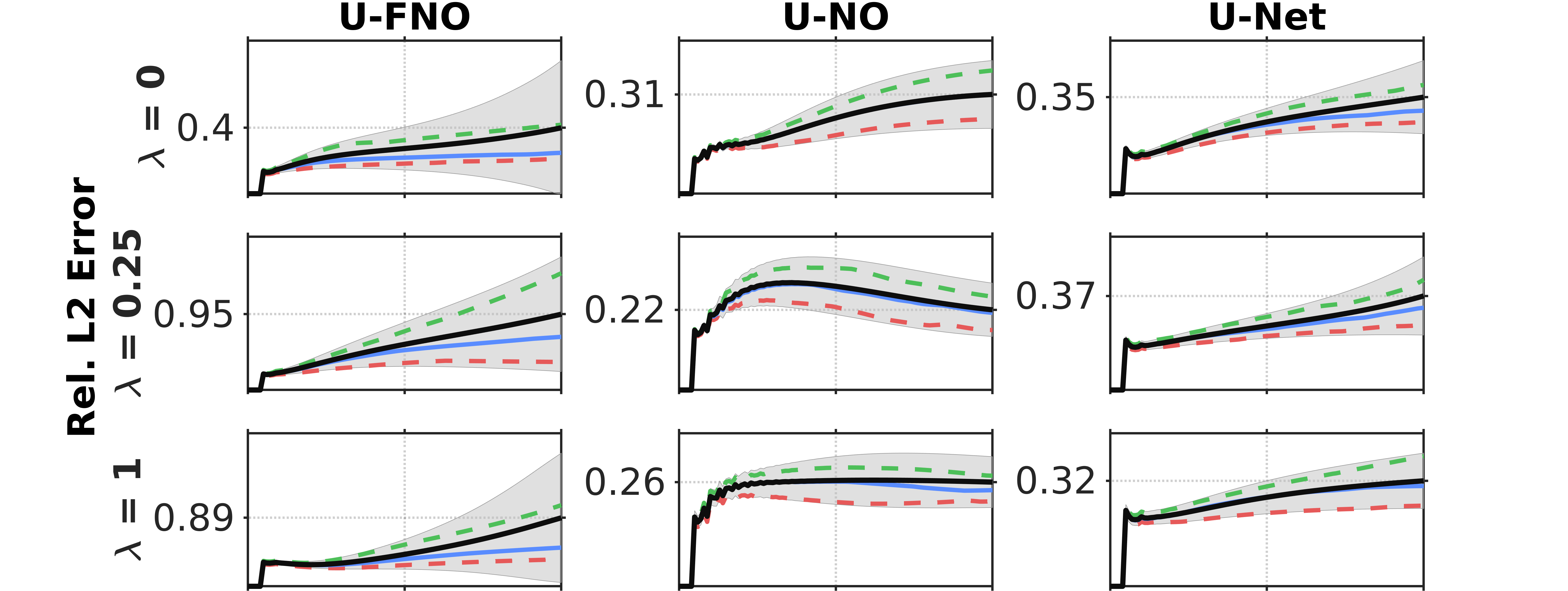}
\end{minipage}%
}

\vspace{0.22em}

\makebox[\textwidth][c]{%
\begin{minipage}[c]{0.025\textwidth}
\centering
\textbf{(b)}
\end{minipage}
\hspace{-0.002\textwidth}
\begin{minipage}[c]{1.025\textwidth}
\centering
\includegraphics[width=\linewidth]{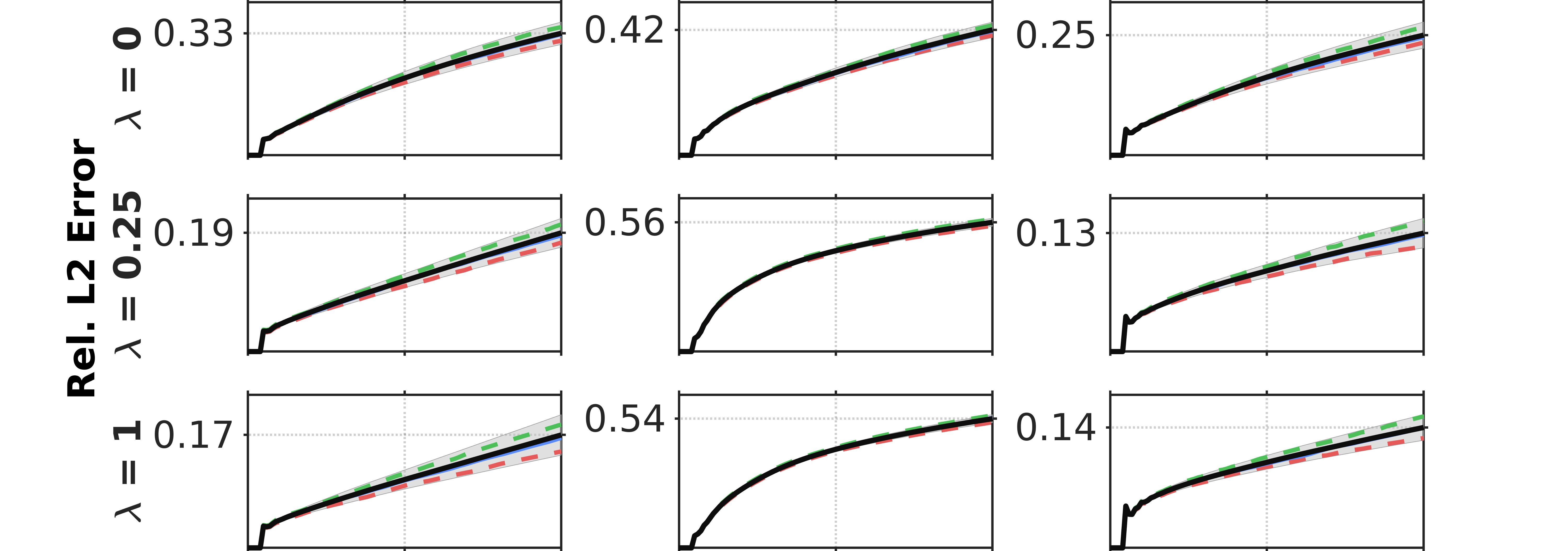}
\end{minipage}%
}

\vspace{0.22em}

\makebox[\textwidth][c]{%
\begin{minipage}[c]{0.025\textwidth}
\centering
\textbf{(c)}
\end{minipage}
\hspace{-0.002\textwidth}
\begin{minipage}[c]{1.025\textwidth}
\centering
\includegraphics[width=\linewidth]{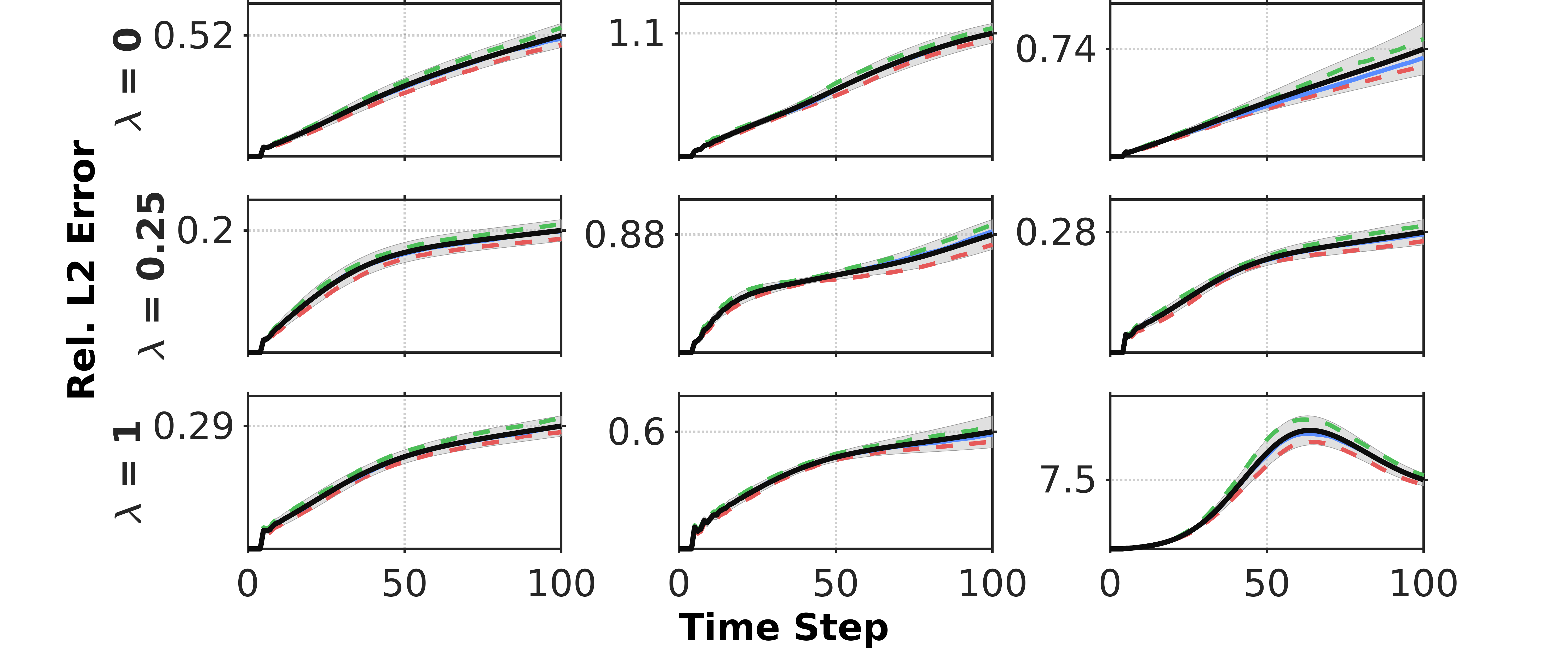}
\end{minipage}%
}

\vspace{-0.15em}

\caption{Rollout relative $L^2$ errors of the U-shaped baselines for the three-dimensional phase-field benchmarks: (a) AC, (b) CH, and (c) SH. The comparison highlights the effect of hierarchical encoder--decoder structure without adaptive local--global fusion.}
\label{fig:ushape_results}
\end{figure*}

\autoref{fig:ushape_results} isolates the U-shaped baselines, namely U-FNO, U-NO, and U-Net. These architectures improve over single-resolution models by using encoder--decoder paths and skip connections, but their performance remains less uniform across the three PDEs because their

\clearpage

\FloatBarrier

\FloatBarrier
\noindent
local and global interactions are either fixed, mainly spectral, or purely convolutional. In contrast, HAMNO uses the same hierarchical principle while replacing fixed feature fusion with adaptive local--global operator selection, which explains its stronger robustness in regimes where the dominant dynamics shift between smooth bulk evolution, sharp interfaces, and long-range interactions.

This comparison also clarifies why multi-scale structure alone is not sufficient. U-Net can recover local spatial details but lacks an explicit global operator mechanism; U-NO provides multi-resolution spectral learning but remains largely governed by global Fourier mixing; and U-FNO combines Fourier and U-Net components through fixed additive coupling. HAMNO strengthens this family of architectures by allowing the model to decide, from the evolving solution itself, how much local or global information should be used at each scale.

\begin{figure}[H]
\centering

\begin{minipage}{0.035\textwidth}
\centering
\textbf{(a)}
\end{minipage}
\hspace{0.005\textwidth}
\begin{minipage}{0.80\textwidth}
\centering
\includegraphics[width=\linewidth]{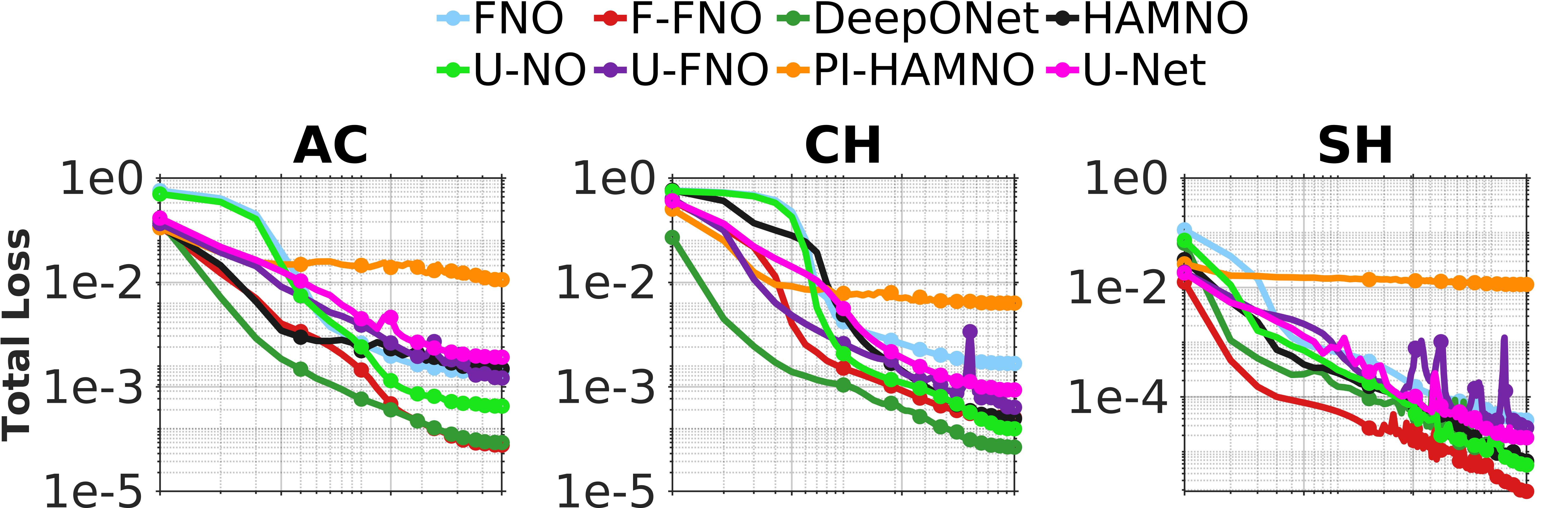}
\end{minipage}

\vspace{0.25em}

\begin{minipage}{0.035\textwidth}
\centering
\textbf{(b)}
\end{minipage}
\hspace{0.005\textwidth}
\begin{minipage}{0.80\textwidth}
\centering
\includegraphics[width=\linewidth]{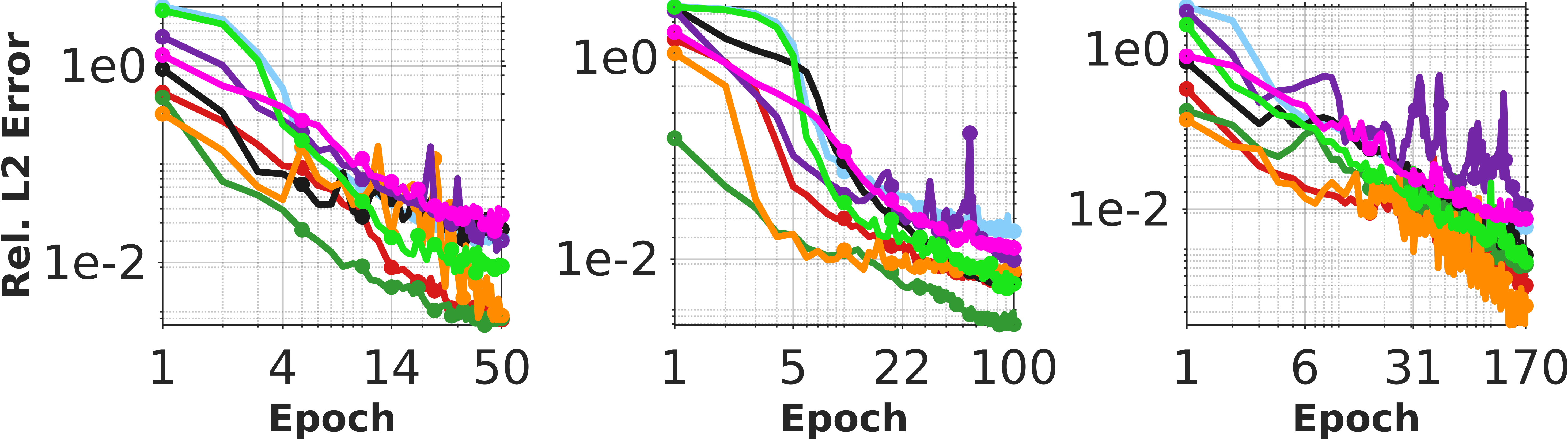}
\end{minipage}

\vspace{-0.25em}

\caption{Training histories of all compared models, showing (a) total loss and (b) test relative $L^2$ error during training.}
\label{fig:summary_training_behavior}
\end{figure}

The training histories in~\autoref{fig:summary_training_behavior}, shown on a logarithmic scale for clarity, indicate that the epoch budgets are sufficient for comparison, with test relative $L^2$ errors reaching the target range for rollout evaluation. The total-loss curves begin at comparable orders of magnitude, showing that the models are optimized from a similar numerical scale and that the later separation is mainly caused by the model structure and the training objective. Data-driven models often show smoother loss decay because they mainly reduce snapshot mismatch. PI-HAMNO may decrease less smoothly because the strong- and weak-form residuals continuously correct the update toward PDE-consistent dynamics, which can temporarily increase the training objective while improving the quality of the learned evolution map.

The test relative $L^2$ curves describe one-step prediction accuracy during training, whereas the previous rollout figures measure repeated self-propagation over the full time horizon. This distinction explains why PI-HAMNO is not always below models such as DeepONet during AC and CH training, although it gives the best long-time rollout accuracy. A lower or smoother epoch-wise one-step error does not necessarily imply a more stable trajectory when the prediction is recursively fed back into the model. PI-HAMNO learns a slightly more constrained update, guided by both the data and the governing residuals, and this produces better error control after many autoregressive steps, especially for the stiffer CH and pattern-forming SH dynamics.

\FloatBarrier

\vspace{0.4em}

\begin{figure*}[!t]
\centering
\includegraphics[width=0.98\textwidth]{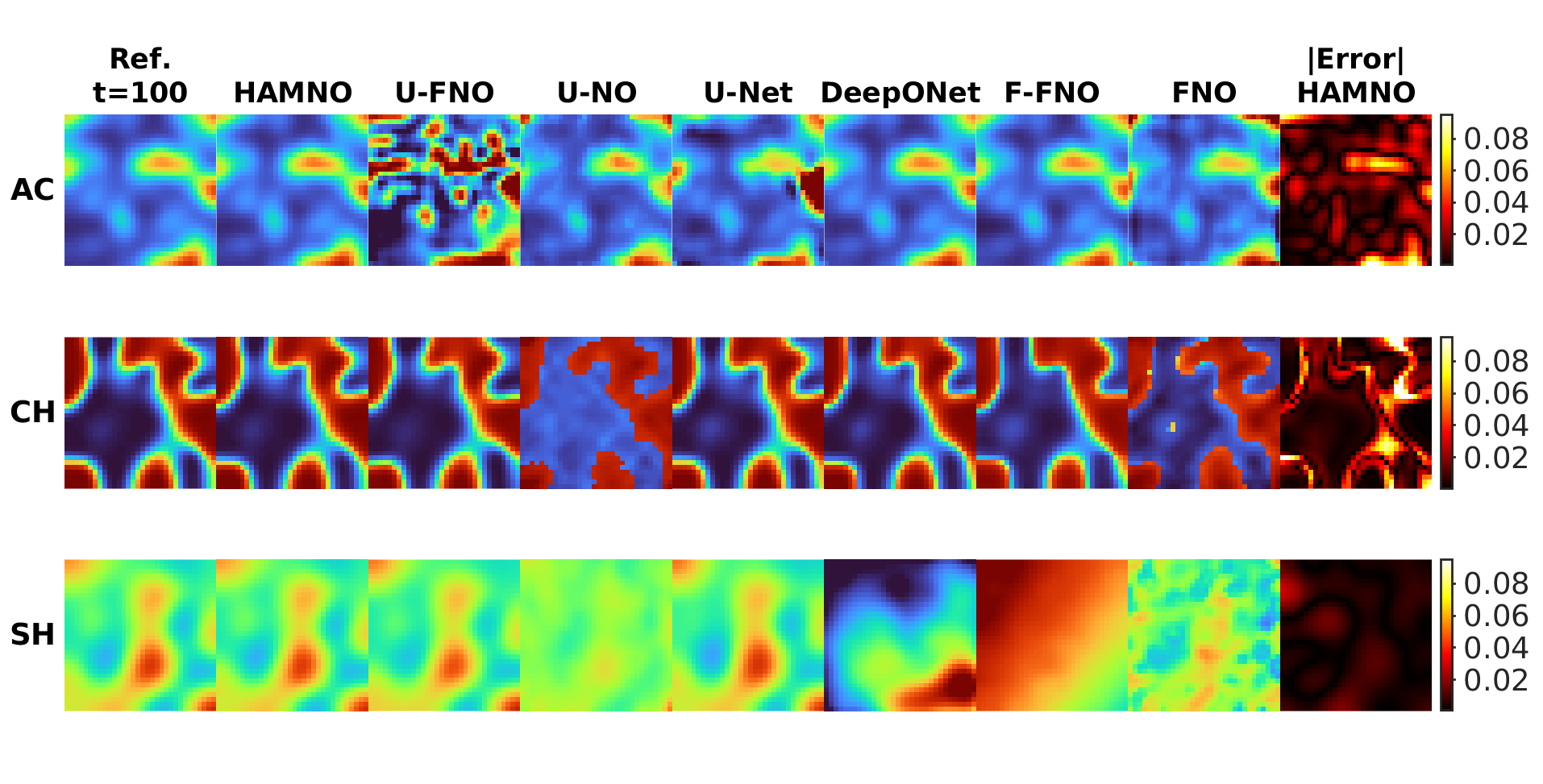}
\vspace{-0.8em}
\caption{Predicted phase-field solutions at $t=100$ for the AC, CH, and SH benchmarks, showing reference fields, model predictions, and the HAMNO absolute error.}
\label{fig:contour_prediction_comparison}
\end{figure*}

A qualitative comparison is shown in~\autoref{fig:contour_prediction_comparison}. The predictions are evaluated at $t=100$ using $100$ training samples and $\lambda=0.25$, where Ref. denotes the numerical solution generated by the DCT-based non-periodic solver. HAMNO accurately reconstructs the dominant morphology of all three phase-field systems, including diffuse interfaces in AC, phase-separated structures in CH, and oscillatory patterns in SH. The remaining discrepancies are mainly localized near sharp transitions and highly oscillatory regions, where even small temporal phase shifts can produce visible pointwise errors during long rollout.

The qualitative results are consistent with the quantitative trends. Methods that rely mainly on spectral representations may capture global organization but can lose localized interfacial detail under non-periodic boundary conditions, while purely local or fixed-fusion hierarchical models can miss longer-range coordination of the evolving field. HAMNO reduces both limitations by combining boundary-sensitive local processing with global operator learning through adaptive fusion, which leads to more coherent long-time reconstructions.

\FloatBarrier

\begin{figure*}[!t]
\centering
\includegraphics[width=0.98\textwidth]{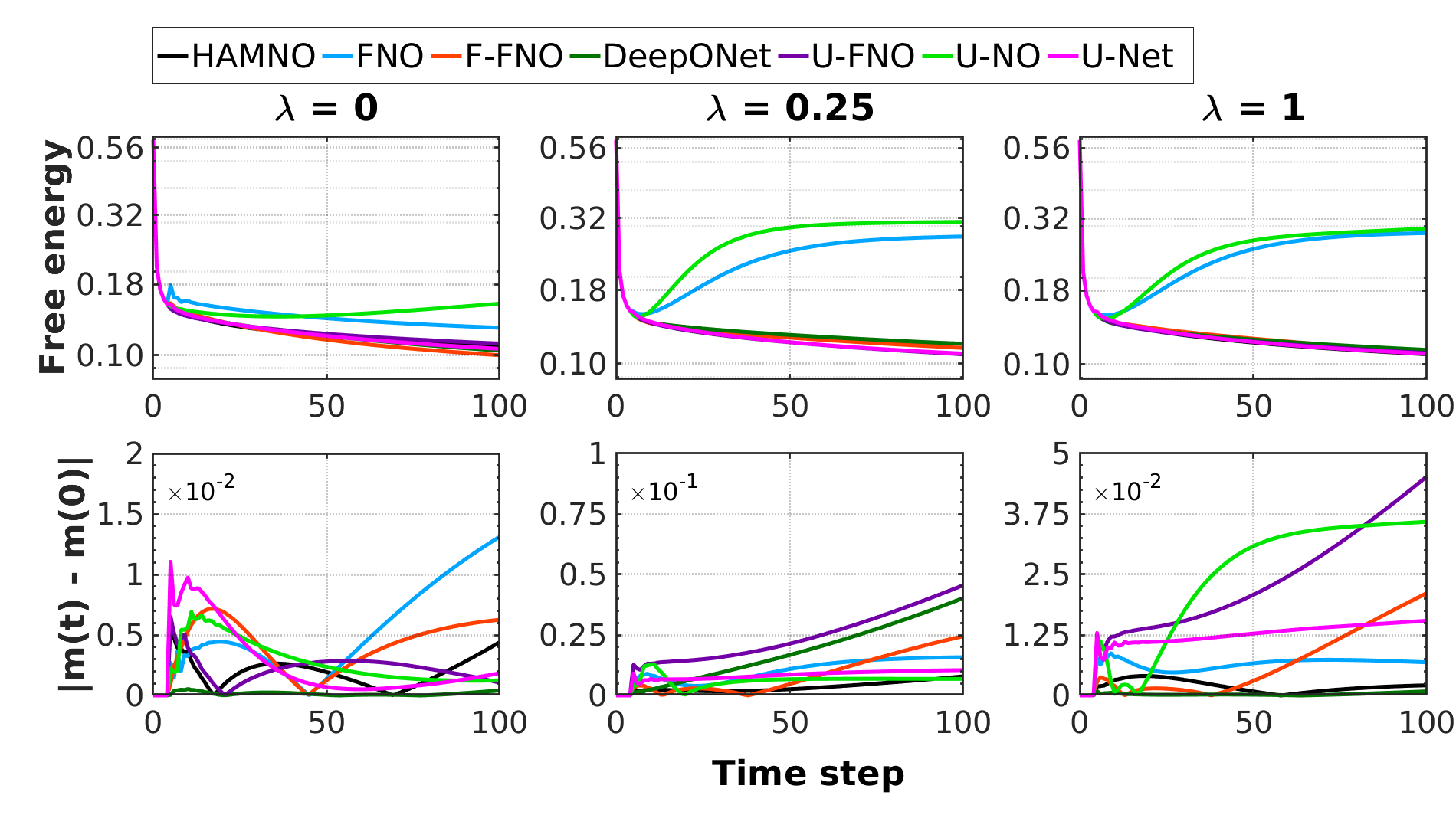}
\vspace{-0.3em}
\caption{Free-energy decay and mass conservation for the CH benchmark under different physics weights $\lambda$.}
\label{fig:ch_energy_mass}
\end{figure*}

The physical diagnostics for the CH benchmark are reported in~\autoref{fig:ch_energy_mass}. The free-energy curves show that HAMNO, DeepONet, U-FNO, F-FNO, and U-Net produce more consistent dissipative behavior, whereas FNO and U-NO deviate more clearly in the physics-informed cases. This behavior suggests that the CH energy evolution benefits from local spatial processing and flexible feature modulation, which are less explicit in FNO, with its mainly global spectral representation, and in U-NO, whose multi-resolution structure is still largely governed by Fourier-based operator blocks. In contrast, DeepONet uses coordinate-dependent modulation of convolutional branch features, U-FNO and F-FNO include local correction paths, U-Net relies on local encoder--decoder features, and HAMNO combines local and global operators adaptively across scales.

The mass plots provide a clearer view of conservation behavior during rollout. In the purely data-driven setting, DeepONet shows very small mass drift, while several other models exhibit visible oscillations, indicating that mass preservation depends strongly on the architecture during autoregressive prediction. In the hybrid case, PI-HAMNO gives the most stable mass behavior, showing that the physics constraint effectively regularizes the learned update when combined with adaptive local--global representation. In the pure-physics setting, HAMNO remains close to the near-zero drift behavior of DeepONet, whereas some baselines still accumulate noticeable deviations. This suggests that mass conservation is improved not only by the physics loss itself, but also by an architecture that can represent both local interfacial motion and global solution balance.

\paragraph{Ablation study of the physics terms.}
To isolate the contribution of the two physics-informed components, we perform an ablation study in the pure-physics setting, $\lambda=1$, by training HAMNO with the strong-form residual, the weak-form residual, and their combined multi-objective formulation. The results are reported in~\autoref{tab:physics_ablation}. For all three benchmarks, the combined strong--weak formulation gives the most stable rollout accuracy, showing that the two residuals provide complementary information rather than redundant constraints.

For the AC equation, the weak form alone is already effective because the variational residual naturally captures the diffusion--reaction balance and handles the Neumann boundary condition in an integral sense. The strong form alone accumulates larger errors at later times, while the combined loss improves long-time stability by using both pointwise PDE consistency and variational smoothing.

\FloatBarrier

For the CH equation, the strong-only formulation performs poorly, indicating that the pointwise mixed residual is not sufficient by itself for this fourth-order mass-conserving dynamics. The weak form substantially improves the result by enforcing the coupled evolution and chemical-potential relations in an integral form, while the full formulation gives the lowest errors by combining local residual accuracy with variational consistency.

For the SH equation, the weak-only case is much less accurate than the strong-only case. This is consistent with the SH formulation used here: the strong residual includes the additional Helmholtz-type auxiliary consistency condition, which provides a stronger constraint for the fourth-order pattern-forming dynamics. The weak form alone enforces the split system variationally, but without this extra consistency it is less able to control the auxiliary field during long rollout. The full loss recovers the best behavior by combining the stabilizing strong-form auxiliary constraint with the weak-form elementwise regularization.

\begin{table}[H]
\centering
\caption{Ablation study of the pure-physics HAMNO setting ($\lambda=1$), where all reported values are mean relative $L^2$ errors at selected rollout times.}
\label{tab:physics_ablation}
\small
\setlength{\tabcolsep}{7pt}
\renewcommand{\arraystretch}{1.14}
\resizebox{\textwidth}{!}{%
\begin{tabular}{lccccccl}
\toprule
Benchmark 
& Strong & Weak 
& $t=20$ & $t=40$ & $t=60$ & $t=80$ & $t=100$ \\
\midrule

AC 
& \checkmark & \checkmark 
& \textbf{5.7308e-02} & \textbf{5.4272e-02} & \textbf{5.5659e-02} & \textbf{5.7625e-02} & \textbf{5.9834e-02} \\
& \checkmark & -- 
& 9.4241e-02 & 1.0352e-01 & 1.0494e-01 & 1.0319e-01 & 1.0080e-01 \\
& -- & \checkmark 
& 6.9575e-02 & 9.9848e-02 & 1.1880e-01 & 1.2869e-01 & 1.3403e-01 \\
\midrule

CH 
& \checkmark & \checkmark 
& \textbf{3.4654e-02} & \textbf{4.2775e-02} & \textbf{4.9068e-02} & \textbf{5.4698e-02} & \textbf{5.9723e-02} \\
& \checkmark & -- 
& 2.1360e+00 & 2.2747e+00 & 2.2382e+00 & 2.2091e+00 & 2.1870e+00 \\
& -- & \checkmark 
& 9.0417e-02 & 1.3160e-01 & 1.6461e-01 & 1.9045e-01 & 2.1091e-01 \\
\midrule

SH 
& \checkmark & \checkmark 
& \textbf{5.7244e-02} & \textbf{7.4171e-02} & \textbf{8.2344e-02} & \textbf{8.5062e-02} & \textbf{8.7741e-02} \\
& \checkmark & -- 
& 8.7098e-02 & 1.2707e-01 & 1.4595e-01 & 1.5430e-01 & 1.6013e-01 \\
& -- & \checkmark 
& 8.7905e-01 & 1.7914e+00 & 2.2084e+00 & 2.3312e+00 & 2.3398e+00 \\
\bottomrule
\end{tabular}%
}
\end{table}

\subsection{Out-of-distribution generalization}

To evaluate whether the learned operators generalize beyond the training distribution, we perform OOD tests using saved models trained on the original datasets. In the ID setting, both training and testing trajectories are generated from the same initial-condition distribution. In the OOD setting, the Gaussian-random-field parameters controlling the initial conditions, denoted by $\alpha$ and $\tau$, are shifted to unseen values or sampled from unseen intervals. This changes the initial length scales and correlation structures of the phase fields while keeping the governing equation, grid resolution, boundary condition, and rollout protocol unchanged.

All OOD evaluations are performed without retraining or fine-tuning. The same autoregressive rollout procedure and relative $L^2$ metric used in the ID experiments are applied, allowing a direct comparison of how each trained model behaves under initial-condition distribution shift. The corresponding results for the AC, CH, and SH benchmarks are reported in~\autoref{tab:ac3d_ood_comparison}, \autoref{tab:ch3d_ood_comparison}, and~\autoref{tab:sh3d_ood_comparison}, respectively.

\begin{table}[H]
\centering
\caption{OOD evaluation results for the 3D AC equation under three unseen initial-condition distributions controlled by $\alpha$ and $\tau$.}
\label{tab:ac3d_ood_comparison}
\resizebox{\textwidth}{!}{%
\begin{tabular}{llccccc}
\hline
\textbf{OOD Setting} &
\textbf{Model} &
\textbf{\# Samples} &
\textbf{Rel. $L^2$ at $t=50$} &
\textbf{Rel. $L^2$ at $t=100$} &
\textbf{Mean Rel. $L^2$} &
\textbf{Sec./Sample} \\
\hline
\multirow{8}{*}{\shortstack{$\alpha=60$ \\ $\tau=260$}}
& PI-HAMNO & 100 & $3.3856 \times 10^{-2}$ & $3.3326 \times 10^{-2}$ & $4.0200 \times 10^{-2}$ & $1.3202$ \\
& HAMNO    & 100 & $9.0570 \times 10^{-2}$ & $1.2771 \times 10^{-1}$ & $8.9824 \times 10^{-2}$ & $1.3554$ \\
& FFNO     & 100 & $9.2818 \times 10^{-2}$ & $1.3605 \times 10^{-1}$ & $8.7555 \times 10^{-2}$ & $1.0646$ \\
& DeepONet & 100 & $1.0473 \times 10^{-1}$ & $1.5275 \times 10^{-1}$ & $9.8308 \times 10^{-2}$ & $3.0392$ \\
& FNO      & 100 & $2.1342 \times 10^{-1}$ & $2.7254 \times 10^{-1}$ & $2.0521 \times 10^{-1}$ & $0.4712$ \\
& U-NO     & 100 & $2.3693 \times 10^{-1}$ & $2.8165 \times 10^{-1}$ & $2.1744 \times 10^{-1}$ & $1.2251$ \\
& U-Net    & 100 & $2.5367 \times 10^{-1}$ & $3.0182 \times 10^{-1}$ & $2.3132 \times 10^{-1}$ & $0.4106$ \\
& U-FNO    & 100 & $2.4549 \times 10^{-1}$ & $3.0289 \times 10^{-1}$ & $2.3000 \times 10^{-1}$ & $1.1188$ \\
\hline
\multirow{8}{*}{\shortstack{$\alpha \in [20,90]$ \\ $\tau \in [160,260]$}}
& PI-HAMNO & 100 & $3.2292 \times 10^{-2}$ & $3.5719 \times 10^{-2}$ & $3.8492 \times 10^{-2}$ & $1.4160$ \\
& HAMNO    & 100 & $9.0278 \times 10^{-2}$ & $1.3199 \times 10^{-1}$ & $8.9339 \times 10^{-2}$ & $1.4058$ \\
& FFNO     & 100 & $1.1520 \times 10^{-1}$ & $1.7759 \times 10^{-1}$ & $1.1066 \times 10^{-1}$ & $1.0856$ \\
& DeepONet & 100 & $1.2649 \times 10^{-1}$ & $2.0046 \times 10^{-1}$ & $1.1999 \times 10^{-1}$ & $3.1267$ \\
& FNO      & 100 & $2.2378 \times 10^{-1}$ & $2.9095 \times 10^{-1}$ & $2.1342 \times 10^{-1}$ & $0.4884$ \\
& U-NO     & 100 & $2.5067 \times 10^{-1}$ & $3.2153 \times 10^{-1}$ & $2.3519 \times 10^{-1}$ & $1.2217$ \\
& U-Net    & 100 & $2.9142 \times 10^{-1}$ & $4.8693 \times 10^{-1}$ & $2.8608 \times 10^{-1}$ & $0.4109$ \\
& U-FNO    & 100 & $3.0111 \times 10^{-1}$ & $5.6823 \times 10^{-1}$ & $3.0729 \times 10^{-1}$ & $0.9952$ \\
\hline
\end{tabular}%
}
\end{table}

\FloatBarrier
\vspace{-0.1cm}

\begin{table}[H]
\centering
\caption{OOD evaluation results for the 3D CH equation under unseen initial-condition distributions controlled by $\alpha$ and $\tau$.}
\label{tab:ch3d_ood_comparison}
\resizebox{\textwidth}{!}{%
\begin{tabular}{llcccccc}
\hline
\textbf{OOD Setting} &
\textbf{Model} &
\textbf{\# Samples} &
\textbf{\shortstack{Rel. $L^2$\\at $t=50$}} &
\textbf{\shortstack{Rel. $L^2$\\at $t=100$}} &
\textbf{\shortstack{Mean Rel.\\$L^2$}} &
\textbf{\shortstack{Mass\\Drift}} &
\textbf{\shortstack{Sec./\\Sample}} \\
\hline

\multirow{8}{*}{\shortstack{$\alpha=20$ \\ $\tau=70$}}
& PI-HAMNO & 100 & $3.7925 \times 10^{-2}$ & $5.0082 \times 10^{-2}$ & $3.6301 \times 10^{-2}$ & $4.0819 \times 10^{-3}$ & $1.2950$ \\
& HAMNO    & 100 & $1.0045 \times 10^{-1}$ & $1.7529 \times 10^{-1}$ & $9.7471 \times 10^{-2}$ & $1.8430 \times 10^{-2}$ & $1.3500$ \\
& DeepONet & 100 & $1.1171 \times 10^{-1}$ & $1.9327 \times 10^{-1}$ & $1.0834 \times 10^{-1}$ & $6.3918 \times 10^{-3}$ & $2.2680$ \\
& U-Net    & 100 & $1.5569 \times 10^{-1}$ & $2.4221 \times 10^{-1}$ & $1.4728 \times 10^{-1}$ & $1.0551 \times 10^{-2}$ & $0.3539$ \\
& FFNO     & 100 & $1.6417 \times 10^{-1}$ & $2.8520 \times 10^{-1}$ & $1.6205 \times 10^{-1}$ & $7.5932 \times 10^{-3}$ & $1.1822$ \\
& U-FNO    & 100 & $2.0426 \times 10^{-1}$ & $3.2553 \times 10^{-1}$ & $1.9173 \times 10^{-1}$ & $1.5355 \times 10^{-2}$ & $1.3690$ \\
& U-NO     & 100 & $2.7453 \times 10^{-1}$ & $4.2029 \times 10^{-1}$ & $2.5742 \times 10^{-1}$ & $3.8185 \times 10^{-3}$ & $1.1735$ \\
& FNO      & 100 & $3.1607 \times 10^{-1}$ & $4.3588 \times 10^{-1}$ & $2.9558 \times 10^{-1}$ & $5.8661 \times 10^{-3}$ & $0.5256$ \\

\hline

\multirow{8}{*}{\shortstack{$\alpha \in [15,35]$ \\ $\tau \in [60,90]$}}
& PI-HAMNO & 100 & $4.2734 \times 10^{-2}$ & $5.8495 \times 10^{-2}$ & $4.0951 \times 10^{-2}$ & $5.5132 \times 10^{-3}$ & $1.2752$ \\
& DeepONet & 100 & $1.1379 \times 10^{-1}$ & $1.9408 \times 10^{-1}$ & $1.0965 \times 10^{-1}$ & $6.0847 \times 10^{-3}$ & $2.2811$ \\
& HAMNO    & 100 & $1.1922 \times 10^{-1}$ & $2.0097 \times 10^{-1}$ & $1.1507 \times 10^{-1}$ & $2.1961 \times 10^{-2}$ & $1.2880$ \\
& U-Net    & 100 & $1.5684 \times 10^{-1}$ & $2.4213 \times 10^{-1}$ & $1.4790 \times 10^{-1}$ & $1.1378 \times 10^{-2}$ & $0.3879$ \\
& FFNO     & 100 & $1.6427 \times 10^{-1}$ & $2.7786 \times 10^{-1}$ & $1.6159 \times 10^{-1}$ & $6.9621 \times 10^{-3}$ & $1.0717$ \\
& U-FNO    & 100 & $2.0484 \times 10^{-1}$ & $3.2315 \times 10^{-1}$ & $1.9194 \times 10^{-1}$ & $1.6566 \times 10^{-2}$ & $0.9718$ \\
& U-NO     & 100 & $2.7393 \times 10^{-1}$ & $4.1443 \times 10^{-1}$ & $2.5759 \times 10^{-1}$ & $4.3455 \times 10^{-3}$ & $1.2220$ \\
& FNO      & 100 & $3.1029 \times 10^{-1}$ & $4.2763 \times 10^{-1}$ & $2.9019 \times 10^{-1}$ & $6.5590 \times 10^{-3}$ & $0.4503$ \\

\hline
\end{tabular}%
}
\end{table}

\begin{table}[H]
\centering
\caption{OOD evaluation results for the 3D SH equation under unseen initial-condition distributions controlled by $\alpha$ and $\tau$.}
\label{tab:sh3d_ood_comparison}
\resizebox{\textwidth}{!}{%
\begin{tabular}{llccccc}
\hline
\textbf{OOD Setting} &
\textbf{Model} &
\textbf{\# Samples} &
\textbf{Rel. $L^2$ at $t=50$} &
\textbf{Rel. $L^2$ at $t=100$} &
\textbf{Mean Rel. $L^2$} &
\textbf{Sec./Sample} \\
\hline

\multirow{8}{*}{\shortstack{$\alpha=220$ \\ $\tau=45$}}
& PI-HAMNO & 100 & $6.1197 \times 10^{-2}$ & $5.6755 \times 10^{-2}$ & $5.1012 \times 10^{-2}$ & $1.2993$ \\
& HAMNO    & 100 & $1.4646 \times 10^{-1}$ & $1.8458 \times 10^{-1}$ & $1.2641 \times 10^{-1}$ & $1.3091$ \\
& U-FNO    & 100 & $2.9764 \times 10^{-1}$ & $5.1954 \times 10^{-1}$ & $2.8013 \times 10^{-1}$ & $0.9420$ \\
& U-Net    & 100 & $3.4900 \times 10^{-1}$ & $6.6067 \times 10^{-1}$ & $3.3556 \times 10^{-1}$ & $0.3874$ \\
& FNO      & 100 & $4.7992 \times 10^{-1}$ & $7.0397 \times 10^{-1}$ & $4.3097 \times 10^{-1}$ & $0.4203$ \\
& U-NO     & 100 & $6.4866 \times 10^{-1}$ & $1.1483 \times 10^{0}$  & $6.2048 \times 10^{-1}$ & $1.2166$ \\
& FFNO     & 100 & $6.1228 \times 10^{-1}$ & $1.8386 \times 10^{0}$  & $7.0430 \times 10^{-1}$ & $1.3111$ \\
& DeepONet & 100 & $1.3422 \times 10^{0}$  & $3.3998 \times 10^{0}$  & $1.3687 \times 10^{0}$  & $2.2778$ \\

\hline

\multirow{8}{*}{\shortstack{$\alpha \in [190,270]$ \\ $\tau \in [30,75]$}}
& PI-HAMNO & 100 & $6.2344 \times 10^{-2}$ & $5.8033 \times 10^{-2}$ & $5.2606 \times 10^{-2}$ & $1.2706$ \\
& HAMNO    & 100 & $1.5948 \times 10^{-1}$ & $1.9734 \times 10^{-1}$ & $1.3904 \times 10^{-1}$ & $1.2664$ \\
& U-FNO    & 100 & $2.9654 \times 10^{-1}$ & $5.2197 \times 10^{-1}$ & $2.8121 \times 10^{-1}$ & $0.9176$ \\
& FNO      & 100 & $4.7666 \times 10^{-1}$ & $6.9830 \times 10^{-1}$ & $4.2877 \times 10^{-1}$ & $0.4117$ \\
& U-Net    & 100 & $3.7256 \times 10^{-1}$ & $7.2875 \times 10^{-1}$ & $3.6173 \times 10^{-1}$ & $0.3580$ \\
& U-NO     & 100 & $6.2346 \times 10^{-1}$ & $1.1130 \times 10^{0}$  & $6.0012 \times 10^{-1}$ & $1.1777$ \\
& FFNO     & 100 & $6.8280 \times 10^{-1}$ & $1.9603 \times 10^{0}$  & $7.8436 \times 10^{-1}$ & $1.3092$ \\
& DeepONet & 100 & $1.2725 \times 10^{0}$  & $3.3256 \times 10^{0}$  & $1.3158 \times 10^{0}$  & $2.2679$ \\

\hline
\end{tabular}%
}
\end{table}

The OOD results show that PI-HAMNO ($\lambda = 0.25$) maintains the strongest predictive accuracy under initial-condition distribution shift across the tested benchmarks. The improvement is most visible at later rollout times, where small one-step errors accumulate and models with weaker inductive bias become less stable. These results indicate that the combination of hierarchical adaptive local--global representation and physics-informed regularization improves not only in-distribution accuracy, but also robustness to unseen initial morphologies.

For CH, the OOD tables also report mass drift. While some baselines can produce small mass drift, their field prediction errors remain substantially larger. This confirms that conservation of a scalar diagnostic alone is not sufficient for accurate OOD trajectory prediction; the full phase-field morphology must also be captured reliably.

\FloatBarrier

\subsection{Robustness with respect to random seed}

To assess whether the observed performance is stable with respect to random initialization and data splitting, we repeat selected experiments using multiple random seeds. First, we compare representative models across the AC, CH, and SH benchmarks using four seeds, $\{0,36,123,1024\}$, with the results reported in~\autoref{tab:seed_robustness_all_models_all_pdes}. This experiment evaluates whether the relative ranking of the models is preserved under moderate randomness in training.

We then perform a more focused robustness study for the proposed model using ten seeds, $\{0,72,123,215,401,2024,4001,8524,10724,24235\}$, as summarized in~\autoref{tab:seed_robustness_hamno_10seeds_all_pdes}. This second test isolates the variability of HAMNO/PI-HAMNO under a broader range of random initializations. For each setting, we report the relative $L^2$ errors at selected rollout times together with the mean, standard deviation, and coefficient of variation across seeds.

\FloatBarrier

\begin{table}[H]
\centering
\caption{Four-seed robustness summary for selected models on the AC, CH, and SH benchmarks, with Mean $\pm$ Std reporting seed statistics of the all-time mean relative $L^2$ error.}
\label{tab:seed_robustness_all_models_all_pdes}
\scriptsize
\setlength{\tabcolsep}{3.2pt}
\renewcommand{\arraystretch}{1.08}
\begin{tabular}{llccclccc}
\toprule
Eq. & Model & $N_{\rm train}$ & $\lambda$ & Tag &
$t=50$ & $t=100$ & Mean $\pm$ Std & Std/Mean \\
\midrule
AC & HAMNO & 100 & 0 & Pure Data & $3.01e-1$ & $5.36e-1$ & $3.08e-1\,\pm\,2.19e-2$ & $0.07$ \\
 &  & 100 & 0.25 & Hybrid & $5.21e-2$ & $5.03e-2$ & $5.29e-2\,\pm\,1.17e-2$ & $0.22$ \\
 &  & 100 & 1 & Pure Phys. & $6.29e-2$ & $7.03e-2$ & $6.70e-2\,\pm\,1.75e-2$ & $0.26$ \\
\addlinespace[1pt]
 & FFNO & 100 & 0 & Pure Data & $1.38e-1$ & $3.21e-1$ & $1.53e-1\,\pm\,4.64e-2$ & $0.30$ \\
 &  & 100 & 0.25 & Hybrid & $4.23e-2$ & $5.11e-2$ & $4.60e-2\,\pm\,7.34e-3$ & $0.16$ \\
 &  & 100 & 1 & Pure Phys. & $6.01e-2$ & $7.75e-2$ & $6.33e-2\,\pm\,8.85e-3$ & $0.14$ \\
\addlinespace[1pt]
 & DeepONet & 100 & 0 & Pure Data & $9.71e-2$ & $1.39e-1$ & $9.16e-2\,\pm\,4.82e-3$ & $0.05$ \\
 &  & 100 & 0.25 & Hybrid & $6.84e-2$ & $7.47e-2$ & $7.69e-2\,\pm\,1.30e-2$ & $0.17$ \\
 &  & 100 & 1 & Pure Phys. & $8.92e-2$ & $9.61e-2$ & $9.58e-2\,\pm\,4.89e-3$ & $0.05$ \\
\addlinespace[1pt]
 & U-Net & 100 & 0 & Pure Data & $3.37e-1$ & $7.69e-1$ & $3.56e-1\,\pm\,1.71e-1$ & $0.48$ \\
 &  & 100 & 0.25 & Hybrid & $2.96e-1$ & $1.28e0$ & $3.97e-1\,\pm\,3.32e-1$ & $0.84$ \\
 &  & 100 & 1 & Pure Phys. & $4.88e-1$ & $1.50e0$ & $5.77e-1\,\pm\,4.35e-1$ & $0.75$ \\
\midrule
CH & HAMNO & 100 & 0 & Pure Data & $1.90e-1$ & $4.53e-1$ & $2.09e-1\,\pm\,1.99e-1$ & $0.95$ \\
 &  & 100 & 0.25 & Hybrid & $6.01e-2$ & $1.30e-1$ & $6.42e-2\,\pm\,3.12e-2$ & $0.49$ \\
 &  & 100 & 1 & Pure Phys. & $6.92e-2$ & $1.63e-1$ & $7.58e-2\,\pm\,3.61e-2$ & $0.48$ \\
\addlinespace[1pt]
 & FFNO & 100 & 0 & Pure Data & $1.70e-1$ & $2.93e-1$ & $1.68e-1\,\pm\,1.03e-2$ & $0.06$ \\
 &  & 100 & 0.25 & Hybrid & $2.15e-1$ & $3.05e-1$ & $2.04e-1\,\pm\,1.03e-1$ & $0.50$ \\
 &  & 100 & 1 & Pure Phys. & $2.06e-1$ & $2.90e-1$ & $1.92e-1\,\pm\,1.26e-1$ & $0.65$ \\
\addlinespace[1pt]
 & DeepONet & 100 & 0 & Pure Data & $1.21e-1$ & $2.05e-1$ & $1.17e-1\,\pm\,7.30e-3$ & $0.06$ \\
 &  & 100 & 0.25 & Hybrid & $1.52e-1$ & $2.24e-1$ & $1.46e-1\,\pm\,6.47e-2$ & $0.44$ \\
 &  & 100 & 1 & Pure Phys. & $1.68e-1$ & $2.41e-1$ & $1.60e-1\,\pm\,7.15e-2$ & $0.45$ \\
\addlinespace[1pt]
 & U-Net & 100 & 0 & Pure Data & $1.82e-1$ & $2.76e-1$ & $1.72e-1\,\pm\,6.69e-3$ & $0.04$ \\
 &  & 100 & 0.25 & Hybrid & $1.12e-1$ & $1.69e-1$ & $1.07e-1\,\pm\,2.06e-2$ & $0.19$ \\
 &  & 100 & 1 & Pure Phys. & $1.32e-1$ & $1.95e-1$ & $1.25e-1\,\pm\,3.12e-2$ & $0.25$ \\
\midrule
SH & HAMNO & 200 & 0 & Pure Data & $1.33e-1$ & $2.13e-1$ & $1.23e-1\,\pm\,2.18e-2$ & $0.18$ \\
 &  & 200 & 0.25 & Hybrid & $5.60e-2$ & $5.36e-2$ & $4.76e-2\,\pm\,5.60e-3$ & $0.12$ \\
 &  & 200 & 1 & Pure Phys. & $8.13e-2$ & $8.63e-2$ & $7.08e-2\,\pm\,2.22e-2$ & $0.31$ \\
\addlinespace[1pt]
 & FFNO & 200 & 0 & Pure Data & $7.18e-1$ & $1.52e0$ & $7.14e-1\,\pm\,3.12e-1$ & $0.44$ \\
 &  & 200 & 0.25 & Hybrid & $3.24e0$ & $1.74e1$ & $4.40e0\,\pm\,4.19e0$ & $0.95$ \\
 &  & 200 & 1 & Pure Phys. & $1.38e0$ & $2.18e0$ & $1.23e0\,\pm\,4.14e-1$ & $0.34$ \\
\addlinespace[1pt]
 & DeepONet & 200 & 0 & Pure Data & $1.16e0$ & $2.99e0$ & $1.27e0\,\pm\,3.21e-1$ & $0.25$ \\
 &  & 200 & 0.25 & Hybrid & $3.28e0$ & $1.34e1$ & $4.47e0\,\pm\,3.49e0$ & $0.78$ \\
 &  & 200 & 1 & Pure Phys. & $3.46e0$ & $3.42e0$ & $2.44e0\,\pm\,3.41e-1$ & $0.14$ \\
\addlinespace[1pt]
 & U-Net & 200 & 0 & Pure Data & $6.82e-1$ & $2.67e0$ & $8.57e-1\,\pm\,9.59e-2$ & $0.11$ \\
 &  & 200 & 0.25 & Hybrid & $5.68e-1$ & $6.89e-1$ & $4.71e-1\,\pm\,3.42e-1$ & $0.73$ \\
 &  & 200 & 1 & Pure Phys. & $3.63e0$ & $5.97e0$ & $3.29e0\,\pm\,2.44e0$ & $0.74$ \\
\bottomrule
\end{tabular}
\end{table}

The seed-robustness results show that the proposed model retains stable long-horizon accuracy across different random seeds. Although some variation is expected due to stochastic initialization and data sampling, the error levels remain within a narrow range relative to the overall performance gap between HAMNO/PI-HAMNO and the weaker baselines. This indicates that the reported improvements are not the result of a favorable single-seed run, but are representative of the model behavior across repeated training trials.

The ten-seed study further confirms that the proposed architecture is not highly sensitive to random initialization. The standard deviations remain modest relative to the mean errors, supporting the reliability of the learned hierarchical local--global representation under repeated training.

\FloatBarrier

\begin{table}[H]
\centering
\caption{Ten-seed robustness summary for HAMNO on the AC, CH, and SH benchmarks, with Mean $\pm$ Std reporting seed statistics of the all-time mean relative $L^2$ error.}
\label{tab:seed_robustness_hamno_10seeds_all_pdes}
\scriptsize
\setlength{\tabcolsep}{3.2pt}
\renewcommand{\arraystretch}{1.08}
\begin{tabular}{llccclccc}
\toprule
Eq. & Model & $N_{\rm train}$ & $\lambda$ & Tag &
$t=50$ & $t=100$ & Mean $\pm$ Std & Std/Mean \\
\midrule
AC & HAMNO & 100 & 0 & Pure Data & $3.24e-1$ & $5.85e-1$ & $3.36e-1\,\pm\,3.98e-2$ & $0.12$ \\
 &  & 100 & 0.25 & Hybrid & $5.14e-2$ & $5.25e-2$ & $5.29e-2\,\pm\,8.61e-3$ & $0.16$ \\
 &  & 100 & 1 & Pure Phys. & $7.44e-2$ & $9.44e-2$ & $7.91e-2\,\pm\,2.77e-2$ & $0.35$ \\
\midrule
CH & HAMNO & 100 & 0 & Pure Data & $2.14e-1$ & $4.30e-1$ & $2.12e-1\,\pm\,2.90e-1$ & $1.37$ \\
 &  & 100 & 0.25 & Hybrid & $5.31e-2$ & $1.06e-1$ & $5.58e-2\,\pm\,2.52e-2$ & $0.45$ \\
 &  & 100 & 1 & Pure Phys. & $5.48e-2$ & $9.51e-2$ & $5.51e-2\,\pm\,1.94e-2$ & $0.35$ \\
\midrule
SH & HAMNO & 200 & 0 & Pure Data & $1.25e-1$ & $2.02e-1$ & $1.16e-1\,\pm\,2.52e-2$ & $0.22$ \\
 &  & 200 & 0.25 & Hybrid & $6.30e-2$ & $5.99e-2$ & $5.28e-2\,\pm\,1.10e-2$ & $0.21$ \\
 &  & 200 & 1 & Pure Phys. & $8.93e-2$ & $9.73e-2$ & $7.73e-2\,\pm\,3.68e-2$ & $0.48$ \\
\bottomrule
\end{tabular}
\end{table}

\FloatBarrier

\section{Conclusion}
\noindent
This work presented HAMNO, a hierarchical adaptive multi-scale neural operator for nonlinear time-dependent systems with non-periodic boundary conditions. By combining local convolutional processing, global spectral interaction, and encoder--decoder multi-scale representation through data-dependent gating, HAMNO adaptively selects the relevant local and global features as the solution evolves.

The in-distribution experiments on the AC, CH, and SH benchmarks demonstrate improved long-horizon rollout accuracy over standard neural-operator baselines, particularly for sharp interfaces, fourth-order dynamics, and pattern-forming regimes. The physics-informed extension, PI-HAMNO, improves stability, physical consistency, and data efficiency by combining data supervision with strong- and weak-form residual constraints. The ablation study shows that this multi-objective strategy is physically meaningful, with the strong form enforcing local PDE consistency and the weak form providing variational regularization that stabilizes the learned dynamics.

The OOD experiments confirm that the trained models retain accurate predictions under unseen initial-condition distributions, indicating robust generalization beyond the training regime. The seed-robustness study further shows that the reported improvements are stable across random initializations and data splits.

Together, these results establish HAMNO and PI-HAMNO as accurate, stable, and data-efficient operator-learning frameworks for non-periodic three-dimensional dynamical systems, while also motivating future extensions to complex geometries and unstructured discretizations through geometry-aware and transformer-based neural operators.

\section*{Acknowledgements}
\noindent
The authors gratefully acknowledge the support of the German Academic Exchange Service (DAAD) through a research scholarship awarded to Mostafa Bamdad at Bauhaus-Universität Weimar. This support contributed to the completion of the work presented in this paper.

\FloatBarrier
\clearpage

\begin{center}
{\Large\bfseries \underline{Supplementary Information}}
\end{center}

\subsection*{Neural-operator hyperparameters}

Table~\ref{tab:nn_hyperparameters} summarizes the training and architectural hyperparameters used for all neural-operator models across the AC, CH, and SH benchmarks. Here, BS denotes the batch size, Exp.\ denotes the expansion factor used in the network blocks, LR is the learning rate, and WD represents the weight-decay regularization parameter. The table also reports the retained spectral modes, latent feature width, input history length $T_{in}$, and the network depth configuration used for each model.

\vspace{0.2em}

\begin{table}[H]
\centering
\scriptsize
\setlength{\tabcolsep}{3pt}
\renewcommand{\arraystretch}{1.18}
\caption{Neural-operator hyperparameters used for the 3D phase-field benchmarks.}
\label{tab:nn_hyperparameters}

\resizebox{\textwidth}{!}{%
\begin{tabular}{llcccccccccc}
\toprule
\textbf{Problem} &
\textbf{Model} &
\textbf{Ep.} &
\textbf{BS} &
\textbf{Exp.} &
\textbf{LR} &
\textbf{WD} &
\textbf{Modes} &
\textbf{Width} &
\textbf{$W_q$} &
\textbf{$T_{in}$} &
\textbf{Depth} \\
\midrule

\multirow{7}{*}{AC3D}
& FNO      & 50  & 8 & 2 & $10^{-3}$ & $5\times10^{-6}$ & 8  & 12 & 12 & 5 & 2 \\
& F-FNO    & 50  & 8 & 2 & $10^{-3}$ & $5\times10^{-6}$ & 12 & 12 & 12 & 5 & 4 \\
& DeepONet & 50  & 8 & -- & $10^{-3}$ & $5\times10^{-6}$ & -- & 64 & 64 & 5 & B4/T4 \\
& U-NO     & 50  & 8 & 2 & $10^{-3}$ & $5\times10^{-6}$ & 8  & 12 & 12 & 5 & 2 \\
& U-Net    & 50  & 8 & -- & $10^{-3}$ & $5\times10^{-6}$ & -- & 12 & -- & 5 & 4-stage \\
& U-FNO    & 50  & 8 & -- & $10^{-3}$ & $5\times10^{-6}$ & 8  & 12 & -- & 5 & 6-block \\
& HAMNO    & 50  & 8 & 2 & $10^{-2}$ & $10^{-3}$         & 8  & 12 & 12 & 5 & 2 \\

\midrule

\multirow{7}{*}{CH3D}
& FNO      & 100 & 8 & 2 & $5\times10^{-4}$ & $5\times10^{-6}$ & 8  & 12 & 12 & 5 & 2 \\
& F-FNO    & 100 & 8 & 2 & $5\times10^{-4}$ & $5\times10^{-6}$ & 12 & 12 & 12 & 5 & 4 \\
& DeepONet & 100 & 8 & -- & $5\times10^{-4}$ & $5\times10^{-6}$ & -- & 64 & 64 & 5 & B4/T4 \\
& U-NO     & 100 & 8 & 2 & $5\times10^{-4}$ & $5\times10^{-6}$ & 8  & 12 & 12 & 5 & 2 \\
& U-Net    & 100 & 8 & -- & $5\times10^{-4}$ & $5\times10^{-6}$ & -- & 12 & -- & 5 & 4-stage \\
& U-FNO    & 100 & 8 & -- & $5\times10^{-4}$ & $5\times10^{-6}$ & 8  & 12 & -- & 5 & 6-block \\
& HAMNO    & 100 & 8 & 2 & $5\times10^{-4}$ & $5\times10^{-6}$ & 8  & 12 & 12 & 5 & 2 \\

\midrule

\multirow{7}{*}{SH3D}
& FNO      & 170 & 8 & 2 & $10^{-3}$ & $5\times10^{-6}$ & 8  & 12 & 12 & 5 & 2 \\
& F-FNO    & 170 & 8 & 6 & $10^{-3}$ & $5\times10^{-6}$ & 12 & 24 & 12 & 5 & 4 \\
& DeepONet & 170 & 8 & -- & $10^{-3}$ & $5\times10^{-6}$ & -- & 64 & 64 & 5 & B4/T4 \\
& U-NO     & 170 & 8 & 2 & $10^{-3}$ & $5\times10^{-6}$ & 8  & 12 & 12 & 5 & 2 \\
& U-Net    & 170 & 8 & -- & $10^{-3}$ & $5\times10^{-6}$ & -- & 12 & -- & 5 & 4-stage \\
& U-FNO    & 170 & 8 & -- & $10^{-3}$ & $5\times10^{-6}$ & 8  & 12 & -- & 5 & 6-block \\
& HAMNO    & 170 & 8 & 2 & $10^{-3}$ & $5\times10^{-6}$ & 8  & 12 & 12 & 5 & 2 \\

\bottomrule
\end{tabular}%
}
\end{table}


\vspace{1.0em}

\subsection*{Rollout accuracy and computational cost}

Tables~\ref{tab:ac3d_parametric_results}--\ref{tab:sh3d_parametric_results} quantify the trends observed in the rollout figures and show how they translate into final-time accuracy and computational cost. 
For AC and CH, the numerical values confirm that the visual reduction in error growth is not limited to a few selected trajectories: HAMNO remains among the most accurate models at $t=100$, and the hybrid PI-HAMNO setting gives the strongest performance in the most data-limited and interfacial regimes.

The SH benchmark gives a sharper test of the same behavior. 
Here, the gap between methods becomes larger, and the table shows that architectures that appear competitive at early times can lose accuracy rapidly as phase errors accumulate. 
HAMNO maintains lower long-time errors across the reported settings, while the additional cost remains the expected price of using hierarchical local--global processing and physics residuals rather than a single spectral path.

\vspace{1em}

\begin{table}[H]
\centering
\caption{Allen--Cahn (3D) rollout errors and computational cost.}
\label{tab:ac3d_parametric_results}
\small
\setlength{\tabcolsep}{5pt}
\renewcommand{\arraystretch}{1.12}
\resizebox{\textwidth}{!}{%
\begin{tabular}{ccclcccc}
\toprule
$N_{\mathrm{train}}$ & $\lambda$ & Tag & Method 
& Rel.\ $L^2$ @ $t=50$ 
& Rel.\ $L^2$ @ $t=100$ 
& Train (s) 
& Infer (s) \\
\midrule
50 & 0 & Pure Data & FNO & 3.1537e-01 & 4.6435e-01 & 20.06 & 0.4304 \\
   &   &           & F-FNO & 1.5221e-01 & 2.7891e-01 & 52.86 & 1.6857 \\
   &   &           & DeepONet & 1.2377e-01 & 1.6647e-01 & 52.16 & 1.1335 \\
   &   &           & \textbf{HAMNO} & \textbf{1.3604e-01} & \textbf{2.4392e-01} & 90.48 & 2.4845 \\
\cmidrule(lr){2-8}
   & 0.25 & Hybrid & FNO & 2.4041e-01 & 2.9039e-01 & 34.51 & 0.6532 \\
   &      &        & F-FNO & 6.9923e-02 & 8.3880e-02 & 85.43 & 1.6574 \\
   &      &        & DeepONet & 7.7465e-02 & 9.2737e-02 & 75.54 & 1.7031 \\
   &      &        & \textbf{HAMNO} & \textbf{8.8918e-02} & \textbf{9.1278e-02} & 79.36 & 2.0526 \\
\cmidrule(lr){2-8}
   & 1.0 & Pure Phys. & FNO & 2.4293e-01 & 2.7478e-01 & 27.85 & 0.4426 \\
   &     &            & F-FNO & 6.3913e-02 & 7.6734e-02 & 69.63 & 1.7126 \\
   &     &            & DeepONet & 7.3443e-02 & 7.9774e-02 & 84.82 & 2.3704 \\
   &     &            & \textbf{HAMNO} & \textbf{5.4844e-02} & \textbf{5.9834e-02} & 101.72 & 2.0976 \\
\midrule
100 & 0 & Pure Data & FNO & 2.2467e-01 & 2.9233e-01 & 33.34 & 0.5448 \\
    &   &           & F-FNO & 9.8792e-02 & 1.5678e-01 & 97.29 & 1.7440 \\
    &   &           & DeepONet & 9.6789e-02 & 1.3943e-01 & 96.43 & 1.1376 \\
    &   &           & \textbf{HAMNO} & \textbf{9.0660e-02} & \textbf{1.3176e-01} & 103.39 & 2.0185 \\
\cmidrule(lr){2-8}
    & 0.25 & Hybrid & FNO & 2.2062e-01 & 2.3194e-01 & 45.47 & 0.5179 \\
    &      &        & F-FNO & 5.6584e-02 & 6.4458e-02 & 135.82 & 3.0271 \\
    &      &        & DeepONet & 4.9329e-02 & 5.3365e-02 & 117.05 & 1.7608 \\
    &      &        & \textbf{HAMNO} & \textbf{4.1246e-02} & \textbf{4.1924e-02} & 133.88 & 2.0419 \\
\cmidrule(lr){2-8}
    & 1.0 & Pure Phys. & FNO & 2.4293e-01 & 2.7478e-01 & 26.90 & 0.4521 \\
    &     &            & F-FNO & 6.3913e-02 & 7.6734e-02 & 78.94 & 1.6510 \\
    &     &            & DeepONet & 7.3443e-02 & 7.9774e-02 & 90.32 & 2.0546 \\
    &     &            & \textbf{HAMNO} & \textbf{5.4844e-02} & \textbf{5.9834e-02} & 78.40 & 2.0248 \\
\bottomrule
\end{tabular}%
}
\end{table}

\vspace{1em}

\begin{table}[H]
\centering
\caption{Cahn--Hilliard (3D) rollout errors and computational cost.}
\label{tab:ch3d_parametric_results}
\small
\setlength{\tabcolsep}{5pt}
\renewcommand{\arraystretch}{1.12}
\resizebox{\textwidth}{!}{%
\begin{tabular}{ccclcccc}
\toprule
$N_{\mathrm{train}}$ & $\lambda$ & Tag & Method 
& Rel.\ $L^2$ @ $t=50$ 
& Rel.\ $L^2$ @ $t=100$ 
& Train (s) 
& Infer (s) \\
\midrule
50 & 0 & Pure Data & FNO & 3.52e-01 & 4.79e-01 & 58.28 & 0.3150 \\
   &   &           & FFNO & 1.91e-01 & 3.25e-01 & 106.21 & 1.5145 \\
   &   &           & DeepONet & 1.43e-01 & 2.47e-01 & 102.79 & 1.1285 \\
   &   &           & \textbf{HAMNO} & \textbf{1.91e-01} & \textbf{2.90e-01} & 113.54 & 1.8595 \\
\cmidrule(lr){2-8}
   & 0.25 & Hybrid & FNO & 4.13e-01 & 5.25e-01 & 192.55 & 0.3180 \\
   &      &        & FFNO & 1.30e-01 & 1.89e-01 & 342.32 & 1.5464 \\
   &      &        & DeepONet & 9.91e-02 & 1.31e-01 & 301.62 & 1.1260 \\
   &      &        & \textbf{HAMNO} & \textbf{4.75e-02} & \textbf{6.82e-02} & 365.31 & 1.9128 \\
\cmidrule(lr){2-8}
   & 1.0 & Pure Phys. & FNO & 4.21e-01 & 5.42e-01 & 197.08 & 0.3217 \\
   &     &            & FFNO & 1.30e-01 & 1.83e-01 & 348.41 & 1.5630 \\
   &     &            & DeepONet & 9.34e-02 & 1.18e-01 & 321.26 & 1.1630 \\
   &     &            & \textbf{HAMNO} & \textbf{4.61e-02} & \textbf{5.97e-02} & 369.44 & 1.8227 \\
\midrule
100 & 0 & Pure Data & FNO & 3.14e-01 & 4.30e-01 & 104.67 & 0.3174 \\
    &   &           & FFNO & 1.70e-01 & 2.90e-01 & 187.05 & 1.5477 \\
    &   &           & DeepONet & 1.17e-01 & 2.02e-01 & 174.09 & 1.1613 \\
    &   &           & \textbf{HAMNO} & \textbf{1.04e-01} & \textbf{1.81e-01} & 198.52 & 1.8736 \\
\cmidrule(lr){2-8}
    & 0.25 & Hybrid & FNO & 4.16e-01 & 5.32e-01 & 332.64 & 0.3110 \\
    &      &        & FFNO & 1.60e-01 & 2.07e-01 & 615.76 & 1.5893 \\
    &      &        & DeepONet & 1.21e-01 & 1.71e-01 & 535.36 & 1.1646 \\
    &      &        & \textbf{HAMNO} & \textbf{3.88e-02} & \textbf{5.03e-02} & 638.14 & 1.9062 \\
\cmidrule(lr){2-8}
    & 1.0 & Pure Phys. & FNO & 4.21e-01 & 5.42e-01 & 194.82 & 0.3148 \\
    &     &            & FFNO & 1.30e-01 & 1.83e-01 & 346.52 & 1.5454 \\
    &     &            & DeepONet & 9.34e-02 & 1.18e-01 & 320.95 & 1.1623 \\
    &     &            & \textbf{HAMNO} & \textbf{4.61e-02} & \textbf{5.97e-02} & 368.67 & 1.8738 \\
\bottomrule
\end{tabular}%
}
\end{table}

\FloatBarrier
\vspace{1cm}

\begin{table}[H]
\centering
\caption{Swift--Hohenberg (3D) rollout errors and computational cost.}
\label{tab:sh3d_parametric_results}
\small
\setlength{\tabcolsep}{5pt}
\renewcommand{\arraystretch}{1.12}
\resizebox{\textwidth}{!}{%
\begin{tabular}{ccclcccc}
\toprule
$N_{\mathrm{train}}$ & $\lambda$ & Tag & Method 
& Rel.\ $L^2$ @ $t=50$ 
& Rel.\ $L^2$ @ $t=100$ 
& Train (s) 
& Infer (s) \\
\midrule
50 & 0 & Pure Data & FNO & 6.2610e-01 & 8.7438e-01 & 94.77 & 0.5440 \\
   &   &           & F-FNO & 4.8784e-01 & 8.0963e-01 & 186.35 & 1.5008 \\
   &   &           & DeepONet & 1.1332e+00 & 2.0494e+00 & 127.37 & 1.1334 \\
   &   &           & \textbf{HAMNO} & \textbf{2.1877e-01} & \textbf{2.8631e-01} & 146.58 & 1.8536 \\
\cmidrule(lr){2-8}
   & 0.25 & Hybrid & FNO & 5.3241e-01 & 6.5465e-01 & 203.20 & 0.4324 \\
   &      &        & F-FNO & 1.0416e+00 & 1.8032e+00 & 557.56 & 1.5406 \\
   &      &        & DeepONet & 3.1406e+00 & 3.9343e+00 & 463.44 & 1.2245 \\
   &      &        & \textbf{HAMNO} & \textbf{9.0603e-02} & \textbf{9.2687e-02} & 456.24 & 1.8790 \\
\cmidrule(lr){2-8}
   & 1.0 & Pure Phys. & FNO & 5.6998e-01 & 6.6208e-01 & 454.02 & 0.7903 \\
   &     &            & F-FNO & 9.9514e-01 & 2.1998e+00 & 592.77 & 1.5441 \\
   &     &            & DeepONet & 4.5384e+00 & 3.2499e+00 & 489.15 & 1.2122 \\
   &     &            & \textbf{HAMNO} & \textbf{7.9495e-02} & \textbf{8.7741e-02} & 455.62 & 1.8954 \\
\midrule
200 & 0 & Pure Data & FNO & 4.8285e-01 & 7.0895e-01 & 355.43 & 0.6222 \\
    &   &           & F-FNO & 6.1783e-01 & 2.0403e+00 & 1078.33 & 1.6161 \\
    &   &           & DeepONet & 1.2063e+00 & 3.2141e+00 & 369.23 & 1.1756 \\
    &   &           & \textbf{HAMNO} & \textbf{1.4331e-01} & \textbf{1.9139e-01} & 451.69 & 1.8563 \\
\cmidrule(lr){2-8}
    & 0.25 & Hybrid & FNO & 1.1206e+00 & 1.1075e+00 & 645.68 & 0.4470 \\
    &      &        & F-FNO & 8.1984e+00 & 2.2923e+01 & 1815.27 & 1.5330 \\
    &      &        & DeepONet & 2.0557e+00 & 4.3374e+00 & 1770.41 & 1.2766 \\
    &      &        & \textbf{HAMNO} & \textbf{5.5386e-02} & \textbf{4.9574e-02} & 1408.64 & 1.8750 \\
\cmidrule(lr){2-8}
    & 1.0 & Pure Phys. & FNO & 5.6998e-01 & 6.6208e-01 & 230.02 & 0.5035 \\
    &     &            & F-FNO & 9.9514e-01 & 2.1998e+00 & 588.50 & 1.5767 \\
    &     &            & DeepONet & 4.5384e+00 & 3.2499e+00 & 481.34 & 1.4554 \\
    &     &            & \textbf{HAMNO} & \textbf{7.9495e-02} & \textbf{8.7741e-02} & 459.24 & 1.8982 \\
\bottomrule
\end{tabular}%
}
\end{table}

\FloatBarrier
\vspace{0.5cm}


\vspace{1.0cm}

\begin{center}
{\Large\bfseries \underline{Appendix}}
\end{center}

\appendix

\section{Governing models and non-periodic dataset generation}
\label{app:dataset_generation}

\subsection{Phase-field models and computational setting}

All datasets are generated on the cubic domain
\begin{equation}
\Omega=\left[-\frac{L}{2},\frac{L}{2}\right]^3,
\end{equation}
using a uniform Cartesian grid with
\begin{equation}
N_x=N_y=N_z=32.
\end{equation}

The phase-field variable is denoted by $u(\mathbf{x},t)$, where
$\mathbf{x}=(x,y,z)\in\Omega$. Homogeneous Neumann boundary conditions are imposed on $\partial\Omega$, such that the normal derivative vanishes at the boundary. This non-periodic setting naturally leads to cosine eigenfunctions and discrete cosine transform (DCT)-based spectral solvers. The governing equations, boundary conditions, and dataset parameters for the AC, CH, and SH systems are summarized in Table~\ref{tab:phasefield_summary}, where $\varepsilon$ denotes the model parameter controlling the interface or pattern scale, $N_t$ is the number of time steps, and $T$ is the final simulation time. The initial conditions are generated from Gaussian random fields parameterized by $\tau$ and $\alpha$, which control the characteristic spatial correlation and morphology of the generated structures.

\begin{table}[H]
\centering
\caption{Summary of the governing phase-field models and dataset parameters.}
\label{tab:phasefield_summary}
\small
\renewcommand{\arraystretch}{1.45}
\setlength{\tabcolsep}{5pt}
\begin{tabular}{p{2.6cm} p{7.0cm} p{3.0cm} p{3.0cm}}
\toprule
\textbf{Model} & \textbf{Governing equation} & \textbf{Boundary condition} & \textbf{Dataset parameters} \\
\midrule

AC
&
Non-conserved gradient flow:
\[
\partial_t u=\Delta u-\frac{1}{\varepsilon^2}(u^3-u)
\]
&
\[
\nabla u\cdot\mathbf{n}=0
\quad \text{on } \partial\Omega
\]
&
\[
\begin{aligned}
L&=2,\\
\varepsilon&=0.1,\\
\Delta t&=10^{-4},\\
N_t&=100,\\
T&=0.01,\\
\tau&\in[290,320],\\
\alpha&\in[90,120]
\end{aligned}
\]

\\
\midrule

CH
&
Conserved phase-separation model:
\[
\partial_t u=\Delta\mu,\qquad
\mu=-\varepsilon^2\Delta u+(u^3-u)
\]
Equivalent form:
\[
\partial_t u=\Delta(u^3-u)-\varepsilon^2\Delta^2u
\]
Semi-implicit split form:
\[
\partial_t u=
2\Delta u-\varepsilon^2\Delta^2u+\Delta(u^3-3u)
\]
&
\[
\begin{aligned}
\nabla u\cdot\mathbf{n}&=0,\\
\nabla\mu\cdot\mathbf{n}&=0
\quad \text{on } \partial\Omega
\end{aligned}
\]
&
\[
\begin{aligned}
L&=2,\\
\varepsilon&=0.05,\\
\Delta t&=10^{-3},\\
N_t&=100,\\
T&=0.1,\\
\tau&=60,\\
\alpha&=15
\end{aligned}
\]

\\
\midrule

SH
&
Fourth-order pattern-forming model:
\[
\partial_t u=\varepsilon u-u^3-(1+\Delta)^2u
\]
Expanded form:
\[
\partial_t u=(\varepsilon-1)u-u^3-2\Delta u-\Delta^2u
\]
&
\[
\nabla u\cdot\mathbf{n}=0
\quad \text{on } \partial\Omega
\]
&
\[
\begin{aligned}
L&=15,\\
\varepsilon&=0.15,\\
\Delta t&=0.05,\\
N_t&=100,\\
T&=5,\\
\tau&\in[280,320],\\
\alpha&\in[80,120]
\end{aligned}
\]

\\

\bottomrule
\end{tabular}
\end{table}

\subsection{DCT-based non-periodic solvers}

For homogeneous Neumann boundary conditions, the Laplacian is diagonalized by cosine modes. For mode indices $p,q,r$, the non-negative Laplacian symbol is
\begin{equation}
\lambda_{pqr}
=
\left(\frac{p\pi}{L_x}\right)^2
+
\left(\frac{q\pi}{L_y}\right)^2
+
\left(\frac{r\pi}{L_z}\right)^2 .
\label{app:eq_cosine_symbol}
\end{equation}
Thus, if $\widehat{u}$ denotes the 3D DCT of $u$, then
\begin{equation}
\widehat{\Delta u}=-\lambda_{pqr}\widehat{u}.
\end{equation}
This allows the linear stiff terms to be treated implicitly and the nonlinear terms explicitly.

\paragraph{Allen--Cahn update.}
For AC, the semi-implicit DCT update is
\begin{equation}
\widehat{u}^{n+1}
=
\frac{
\widehat{u}^{n}
-
\dfrac{\Delta t}{\varepsilon^2}
\widehat{(u^3-u)}^{\,n}
}
{
1+\Delta t\,\lambda_{pqr}
}.
\label{app:eq_ac_solver}
\end{equation}

\paragraph{Cahn--Hilliard update.}
For CH, the corresponding DCT-based semi-implicit update is given by
\begin{equation}
\widehat{u}^{n+1}
=
\frac{
\widehat{u}^{n}
-
\Delta t\,\lambda_{pqr}\widehat{(u^3-3u)}^{\,n}
}
{
1+\Delta t\left(2\lambda_{pqr}+\varepsilon^2\lambda_{pqr}^2\right)
}.
\label{app:eq_ch_solver}
\end{equation}
The zero-frequency mode is preserved, which is consistent with mass conservation.

\paragraph{Swift--Hohenberg update.}
For SH, the DCT-based semi-implicit update is
\begin{equation}
\widehat{u}^{n+1}
=
\frac{
\widehat{(u^n/\Delta t)}
-
\widehat{(u^3)}^{\,n}
+
2\lambda_{pqr}\widehat{u}^{\,n}
}
{
1/\Delta t+(1-\varepsilon)+\lambda_{pqr}^2
}.
\label{app:eq_sh_solver}
\end{equation}

\subsection{Initial conditions and stored datasets}

For all three problems, initial conditions are generated from three-dimensional Gaussian random fields followed by thresholding into binary phases. For AC and SH, a random shift is applied before thresholding to enrich the initial morphology. For CH, the thresholded field is shifted to have zero discrete mean,
\begin{equation}
u_0
\leftarrow
u_0
-
\frac{1}{|\Omega_h|}
\sum_{\mathbf{x}_i\in\Omega_h}u_0(\mathbf{x}_i),
\end{equation}
so that the initial mass is consistent with the conservative dynamics.

Each dataset contains $600$ trajectories. Every trajectory stores $101$ frames, including the initial condition and $100$ time steps. The raw binary files are converted into MATLAB \texttt{.mat} files and then arranged in the learning pipeline as tensors of the form
\begin{equation}
(N_t,N_x,N_y,N_z),
\end{equation}
which are windowed into input histories and one-step prediction targets. The resulting datasets are then used in the learning pipeline following the numerical phase-field simulation and neural-operator training protocols described in~\cite{yoon2020fourier,BAMDAD2026118862}.

\bibliographystyle{unsrt}
\bibliography{references}

\end{document}